\documentclass[acmsmall, screen]{acmart}

\AtBeginDocument{%
  }

\renewcommand\footnotetextcopyrightpermission[1]{}

\usepackage{ragged2e}
\usepackage{fontawesome}

\usepackage{tabularx}         
\usepackage{amsfonts}         
\usepackage{bbding}           
\usepackage{booktabs}         

\usepackage{amsmath}          
\usepackage{pifont}           
\usepackage{color}

\usepackage{multirow}         

\usepackage{caption}
\usepackage{subcaption}       
\usepackage{enumitem}         
\usepackage{flushend}         
\usepackage{balance}          

\usepackage{lineno}           
\usepackage{algorithm}
\usepackage{algorithmic}
\usepackage{graphicx}
\usepackage{textcomp}
\usepackage{colortbl}         
\usepackage{amsthm}           
\usepackage{url}              
\usepackage{float}            
\usepackage{multicol}
\usepackage{bm}
\usepackage{bbm}
\usepackage{tikz}
\usepackage{forest}

\usepackage[utf8]{inputenc}   
\usepackage{fontenc}          
\usepackage{nicefrac}         
\usepackage{microtype}        
\usepackage{wrapfig}          

\newcommand{\cmark}{\ding{51}}%
\newcommand{\xmark}{\ding{55}}%
\newcommand{\comicsunderstanding}{Comics Understanding}

\newcommand*\colourcheck[1]{%
  \expandafter\newcommand\csname #1check\endcsname{\textcolor{#1}{\ding{52}}}%
}
\colourcheck{blue}
\colourcheck{green}
\colourcheck{red}

\newcommand*\colourwrong[1]{%
  \expandafter\newcommand\csname #1wrong\endcsname{\textcolor{#1}{\ding{55}}}%
}
\colourwrong{blue}
\colourwrong{green}
\colourwrong{red}

\newcommand{\artColor}{\textcolor{betterOrange}{\faPaintBrush}}
\newcommand{\artBW}{\textcolor{gray}{\faPaintBrush}}
\definecolor{lightgray}{HTML}{F3F3F3}
\definecolor{betterOrange}{RGB}{255,140,0}

\usepackage{array}
\newcolumntype{L}[1]{>{\raggedright\let\newline\\\arraybackslash\hspace{0pt}}m{#1}}
\newcolumntype{C}[1]{>{\centering\let\newline  \\\arraybackslash\hspace{0pt}}m{#1}}
\newcolumntype{R}[1]{>{\raggedleft\let\newline \\\arraybackslash\hspace{0pt}}m{#1}}

\begin{document}

\title{One missing piece in Vision and Language: A Survey on Comics Understanding}

\author{Emanuele Vivoli}
\affiliation{
\institution{Computer Vision Center, UAB}
\city{Barcelona}
\country{Spain}
}%
\affiliation{
\institution{MICC, University of Florence}
\city{Florence}
\country{Italy}
}
\email{evivoli@cvc.uab.cat}

\author{Mohamed Ali Souibgui}
\affiliation{%
    \institution{Computer Vision Center, UAB}
    \city{Barcelona}
    \country{Spain}
}

\author{Andrey Barsky}
\affiliation{%
    \institution{Computer Vision Center, UAB}
    \city{Barcelona}
    \country{Spain}
}

\author{Artemis Llabrés}
\affiliation{%
    \institution{Computer Vision Center, UAB}
    \city{Barcelona}
    \country{Spain}
}

\author{Marco Bertini}
\affiliation{%
    \institution{MICC, University of Florence}
    \city{Florence}
    \country{Italy}
}

\author{Dimosthenis Karatzas}
\affiliation{%
    \institution{Computer Vision Center, UAB}
    \city{Barcelona}
    \country{Spain}
}




\begin{abstract}
Vision-language models have recently evolved into versatile systems capable of high performance across various tasks, often in zero-shot settings. Comics Understanding, a complex and multifaceted field, stands to benefit from these advances greatly. As a medium, comics combine rich visual and textual narratives, challenging AI models with tasks that span image classification, object detection, instance segmentation, and deeper narrative comprehension through sequential panels. However, the unique structure of comics—marked by stylistic variation, reading order complexity, and non-linear storytelling—poses distinct challenges. In this survey, we provide a comprehensive review of Comics Understanding from both dataset and task perspectives. Our contributions include: (1) analyzing the unique structure and elements of the comics medium; (2) surveying key datasets and tasks in comics research; (3) introducing the Layer of Comics Understanding (LoCU) framework, a novel taxonomy for redefining vision-language tasks in comics; (4) categorizing existing methods using the LoCU framework; and (5) identifying research challenges and future directions for applying vision-language models to comics. This survey pioneers a task-oriented framework for comics intelligence, aiming to guide future research by addressing gaps in data and task definition. A project associated with this survey is available at \url{https://github.com/emanuelevivoli/awesome-comics-understanding}.
\end{abstract}

\maketitle
\section{{Introduction}}

Comics serve as a sophisticated medium that combines visual and textual elements to tell stories~\cite{mccloud_understanding_1998}. Applying recent machine learning tools to understand comics is complicated. This complexity comes from its unique domain that, driven by the author’s creativity, exhibits unique variations in both style and content. For example, modeling the drawing style is complicated, the reading order changes from one book to another, and, unlike natural images, comics frequently depict transformations that defy physical laws, adding another layer of complexity.

Motivated by these challenges, advanced artificial intelligence (AI) approaches have been increasingly applied to \comicsunderstanding. Given that AI thrives on tackling complex and diverse tasks, many researchers are now focusing on problems like
object detection~\cite{vivoli_cdf_2024}, semantic segmentation~\cite{bhattacharjee_github_2024}, optical character recognition (OCR)~\cite{baek_coo_2022}, recurrence of characters and objects in varying contexts~\cite{li_dual_2020}, pose estimation~\cite{augereau_survey_2018}, depth estimation~\cite{bhattacharjee_estimating_2022}, and integration of visual and textual narratives~\cite{iyyer_amazing_2017}. Comics also include complex narrative elements such as satire and irony and spatial and temporal dynamics within the story that have not been deeply investigated. Many of these tasks, individually or in combination, represent the cutting edge of advanced AI, particularly in the area of Multimodal Vision-Language understanding~\cite{liu_improved_2024,li2024llavanext-strong,xue_xgenmm_2024,laurencon_building_2024,yao_minicpmv_2024}.

\begin{wrapfigure}{r}{0.6\textwidth}
\centering
\includegraphics[width=0.6\textwidth]{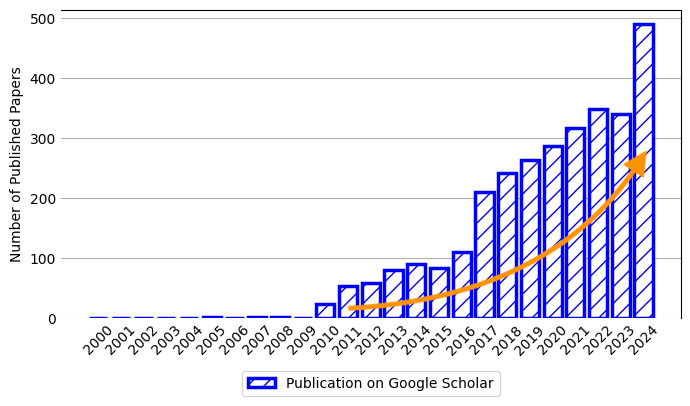}
\vspace{-0.2in}
\caption{Number of publications on Comics and Manga (from Google Scholar). The publications have been filtered by keywords (manga, comics, graphic novels), topics, and journals (computer science).}
\label{fig:trend}
\vspace{-0.1in}
\end{wrapfigure}

Historically, in machine learning (and deep learning), innovation has been driven by the available data, both in terms of tasks and more capable models.
As new types of annotated data and more complex tasks emerged, the need for innovative approaches grew, particularly in domains that combine multiple modalities.
The domain of multimodal learning, especially the convergence of vision and language, has witnessed significant advancements. These advancements include the shift away from independent uni-modal representations \cite{radford_learning_2021,jia_scaling_2021} toward the alignment and integration of different modalities \cite{cho_unifying_2021, li_blip2_2023, singh_flava_2022, zhu_minigpt4_2023, chen_minigptv2_2023, wang_cogvlm_2023} and the creation of multimodal embedding space combining aligned uni-modal inputs \cite{girdhar_imagebind_2023,zhu_languagebind_2024}, with early and late fusion. Trained on extensive multimodal datasets, these models have shown remarkable abilities in tasks like multimodal retrieval \cite{radford_learning_2021,girdhar_imagebind_2023,zhu_languagebind_2024}, visual question answering, image captioning \cite{li_blip2_2023}, reasoning \cite{chen_minigptv2_2023,wang_cogvlm_2023}, and retrieval-augmented multimodal generation \cite{yasunaga_retrievalaugmented_2023} as well as image generation \cite{gal_image_2022,yang_diffusion_2022}.

The comics domain is uniquely well-suited to driving advancements in these types of multimodal reasoning models. Research in comics has extensively explored a range of questions, from the human ability to derive meaning from sequential images \cite{cohn_visual_2013} to machine interpretation of comic strips, particularly through closure tasks \cite{iyyer_amazing_2017}.
With the continuous advancement in AI models and the intensive interest in harvesting vast knowledge from Comics, a lot of work is being published constantly. This is evidenced by a great number of recent papers as shown in Fig.~\ref{fig:trend}.
These contributions shifted the landscape around \comicsunderstanding, leaving previous surveys outdated~\cite{augereau_survey_2018}.
Nowadays, the \comicsunderstanding~ research community lacks a comprehensive survey that catalogs these papers and the various challenges in the Vision-Language domain.

To address this need, we introduce in this paper the novel \textit{Layer of \comicsunderstanding} (LoCU) framework.  
This taxonomy, illustrated in Table \ref{tab:locu}, delineates tasks by their input/output modalities and the spatio-temporal dimensions necessary for reasoning over comic data, inviting a structured and progressive approach to comics analysis.
Motivated by the uniquely broad task distribution of the comics medium, this survey aggregates and examines all extant research on machine learning in comics, aimed at highlighting critical unresolved issues in the field. It is the first to review the breadth of comics research in a hierarchical structure, and to our knowledge, the first to propose a big-picture taxonomy of comics intelligence from a task-oriented perspective.

\begin{figure*}[!t]
    \centering
    \includegraphics[width=\textwidth]{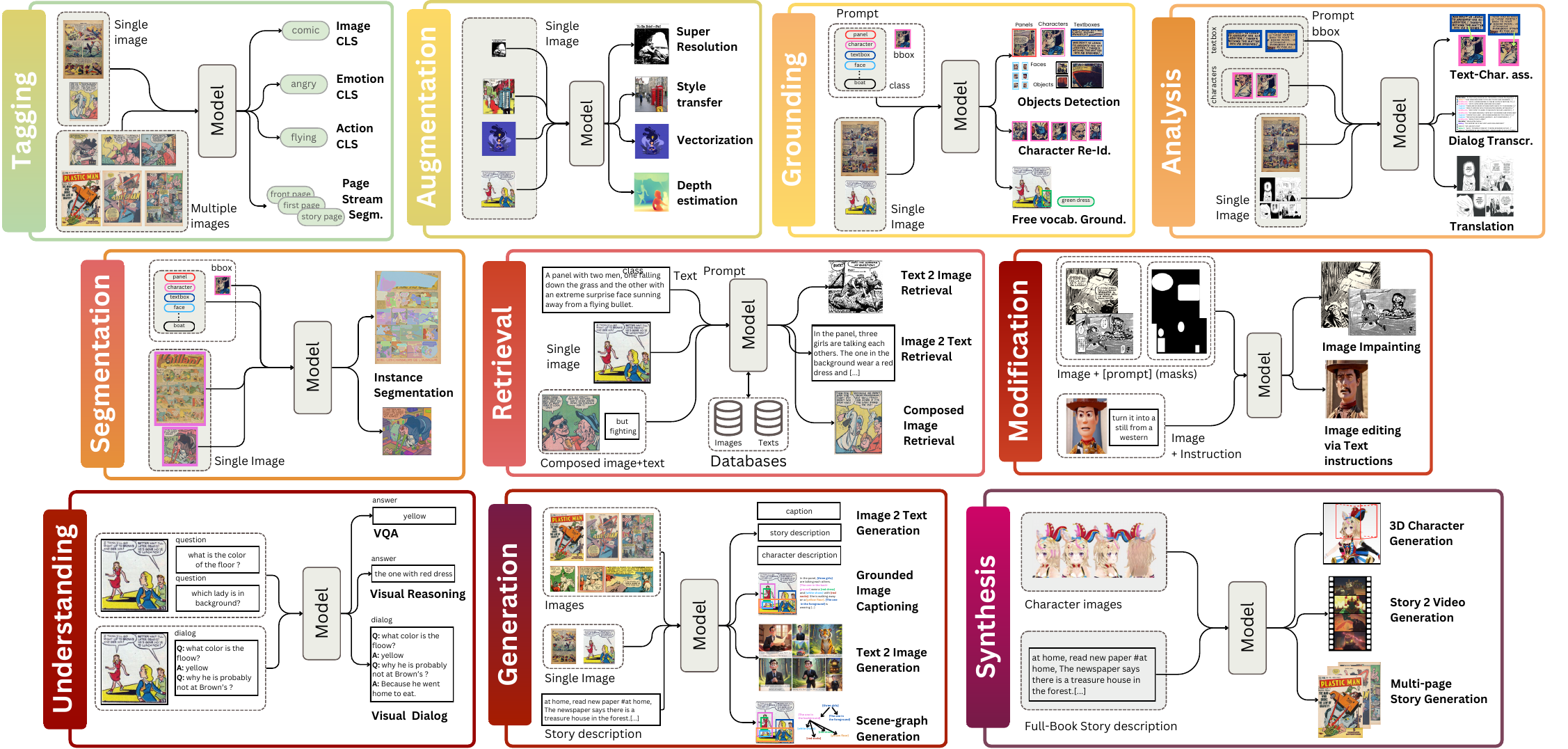}
    \caption{ Visualization of the tasks in Layers of \comicsunderstanding. From panel level to multipage, from unimodal to multimodal, and from simplest to more complex.}
    \label{fig:tasks}
\end{figure*}
%
Within this framework, the survey enumerates the state of the art in \comicsunderstanding~tasks, including image classification, object detection, semantic segmentation, style transfer, multimodal retrieval, synthesis, and generation. Our comprehensive analysis uncovers fundamental gaps—specifically in data availability and task definition.

In summary, the main contributions of this work are threefold. 
    \textit{First}, it presents a systematic review of 
    \comicsunderstanding~tasks. This is the \textit{first} survey of comics tasks for the Vision-Language domain, which provides a big picture of this promising research field with a comprehensive summary and categorization of existing studies.
    \textit{Second}, it studies the up-to-date progress of \comicsunderstanding, including a comprehensive benchmarking and discussion of existing work over multiple public datasets.
    \textit{Third}, it shares several research challenges and potential research directions that could be pursued in VLMs 
    for \comicsunderstanding.

The rest of this survey is organized as follows. 
Section \ref{sec:background} introduces the Comics medium as a composition of Visual-Language tasks, and several related surveys. 
Section \ref{sec:foundation} describes the Foundations of Comics, including their structure, similarities, and differences among various types. 
Section \ref{sec:dataset} introduces existing tasks and the commonly used datasets in \comicsunderstanding, also focusing on evaluation procedures.
In Section \ref{sec:locu} we propose the Layer of \comicsunderstanding~taxonomy (LoCU), and for each layer a collection of tasks grouped by difficulty. In Section \ref{sec:locu-layer1} we start with Layer 1: Tagging and Augmentation, Section \ref{sec:locu-layer2} with Layer 2: Grounding, Analysis, and Segmentation, in Section \ref{sec:locu-layer3} with Layer 3: Retrieval and Modification, Section \ref{sec:locu-layer4} with Layer 4: Understanding, and in Section \ref{sec:locu-layer5} with Layer 5: Generation and Synthesis.
Finally, we conclude with Section \ref{sec:conclusion}.

\section{Background}\label{sec:background}

This section first presents the development of the comics research field based on the learning paradigm and how it evolves towards \textit{more complex tasks}.
Then, we introduce the main limitations encountered by the research community, as well as discuss several related surveys to highlight the scope and contributions of this work.

\subsection{The Comics Research Epochs}

The development of the comics field can be broadly divided into three phases, including (1) \textit{Traditional Machine Learning}, (2) \textit{Deep Learning}, and (3) \textit{Modern Foundational Models}.
In what follows, we introduce, compare, and analyze the three paradigms and how they shaped Comics advancements in detail.
\\
\noindent
\textbf{Traditional Machine Learning.} Early machine learning approaches heavily depended on \textit{feature engineering}, utilizing hand-crafted features~\cite{chou_comic_2011,amirshahi_phog_2012_new,sun_detection_2013}, which were capable of, \textit{e.g.}, classify the style of a manga using SVMs~\cite{chu_linebased_2014_new}, retrieve a manga character from a sketch with edge orientation histograms~\cite{matsui_sketch2manga_2014} or segment comic balloons with energy functions and active contour models~\cite{rigaud_active_2013}. However, these paradigms require domain experts to craft effective features for specific tasks, which do not cope well with complex tasks and have poor generalizability. Moreover, the feature engineering obstacles the extension of the \comicsunderstanding~to more tasks.
\\
\noindent
\textbf{Deep Learning approaches.} With the development of deep learning, researchers have achieved major improvements by using end-to-end trainable deep neural networks (DNNs). These models eliminate the need for complex \textit{feature engineering} and shift the focus to designing network architectures that learn features automatically. Applying these paradigms to comics came as a natural step. Models became better at detection \cite{qin_faster_2017,ogawa_object_2018} and segmentation \cite{dubray_deep_2019,dutta_cnnbased_2020,nguyen_learning_2020}, and more complex tasks started being investigated and solved (e.g. visual and textual closure \cite{iyyer_amazing_2017}).
However, the turn from traditional machine learning toward deep learning raises a new grand challenge: the laborious collection of large-scale, task-specific, and crowd-labeled data in DNN training. Moreover, while deep learning methods excel in specific tasks, they are often difficult to adapt to multi-task. In the domain of comics, where data is scarce, protected by copyright, and challenging to annotate, these constraints have significantly hindered the development of more generalized, multi-task models for \comicsunderstanding.
\\
\noindent
\textbf{Modern Foundational models.} The era of foundational models marks a significant evolution in the application of AI to comics, with models that can learn from vast amounts of unstructured data to perform a variety of complex tasks~\cite{radford_learning_2021,oquab_dinov2_2023,liu_grounding_2023}.
These models relax previous limitations by ``solving'' simple yet important tasks as a zero-shot counter effect, e.g. panel, character, and text-box detection~\cite{sachdeva_manga_2024}. By leveraging these foundational models, more advanced tasks can be proposed and takled~\cite{vivoli_comix_2024,sachdeva_tails_2024}, opening the doors to yet unexplored more complex ones.

\renewcommand\arraystretch{0.9}
\begin{table*}[!t]
\scriptsize
\centering
    \caption{Overview of Vision-Language Tasks of the \textit{Layers of \comicsunderstanding}. The ranking is based on input and output modalities and dimensions, as illustrated in Supplementary Materials.}
    
    \begin{tabular}{ccccc}

\toprule

\rowcolor[HTML]{D9D9D9}  \textbf{Layer} &   \textbf{Category} &   \textbf{Task} &  \textbf{Input} &  \textbf{Output} \\

\midrule
  
\cellcolor[HTML]{6AA84F}{\color[HTML]{FFFFFF} \begin{tabular}[c]{@{}c@{}}0\\ (Sec.5)\end{tabular}} &  \cellcolor[HTML]{6AA84F}{\color[HTML]{FFFFFF} \textbf{\begin{tabular}[c]{@{}c@{}}View\\ (Sec. 5.1)\end{tabular}}} &  \cellcolor[HTML]{B6D7A8}Basic Image Viewing (BIV) &  \cellcolor[HTML]{FFFFFF}Text Command &  \cellcolor[HTML]{FFFFFF}Image Display \\ 

\midrule

\cellcolor[HTML]{DED06E}{\color[HTML]{FFFFFF} } &  \cellcolor[HTML]{B6D7A8}{\color[HTML]{FFFFFF} } &  \cellcolor[HTML]{D9EAD3}Image Classification (I-CLS) &  \cellcolor[HTML]{EFEFEF}Image &  \cellcolor[HTML]{EFEFEF}Tag \\

\cellcolor[HTML]{DED06E}{\color[HTML]{FFFFFF} } &  \cellcolor[HTML]{B6D7A8}{\color[HTML]{FFFFFF} } &  \cellcolor[HTML]{D9EAD3}Emotion Classification &  Comic Panels/Images &  Emotion Labels \\

\cellcolor[HTML]{DED06E}{\color[HTML]{FFFFFF} } &  \cellcolor[HTML]{B6D7A8}{\color[HTML]{FFFFFF} } &  \cellcolor[HTML]{D9EAD3}Action Detection &  \cellcolor[HTML]{EFEFEF}Multiple Panels &  \cellcolor[HTML]{EFEFEF}Tag \\

\cellcolor[HTML]{DED06E}{\color[HTML]{FFFFFF} } &  \multirow{-4}{*}{\cellcolor[HTML]{B6D7A8}{\color[HTML]{FFFFFF} \textbf{\begin{tabular}[c]{@{}c@{}}Tagging\\ (Sec. 6.1)\end{tabular}}}} &  \cellcolor[HTML]{D9EAD3}Page Stream Segmentation (PSS) &  Images &  Tags sequence \\

\cellcolor[HTML]{DED06E}{\color[HTML]{FFFFFF} } &  \cellcolor[HTML]{DDD06E}{\color[HTML]{FFFFFF} } &  \cellcolor[HTML]{FFF2CC} Image Super-Resolution (ISR) &  \cellcolor[HTML]{EFEFEF}Image &  \cellcolor[HTML]{EFEFEF}Image \\

\cellcolor[HTML]{DED06E}{\color[HTML]{FFFFFF} } &  \cellcolor[HTML]{DDD06E}{\color[HTML]{FFFFFF} } &  \cellcolor[HTML]{FFF2CC}Style Transfer (ST) &  \cellcolor[HTML]{FFFFFF}Image &  \cellcolor[HTML]{FFFFFF}Image \\

\cellcolor[HTML]{DED06E}{\color[HTML]{FFFFFF} } &  \cellcolor[HTML]{DDD06E}{\color[HTML]{FFFFFF} } &  \cellcolor[HTML]{FFF2CC}Vectorization &  \cellcolor[HTML]{EFEFEF}Comic Panels/Images &  \cellcolor[HTML]{EFEFEF}Vector Image \\

\multirow{-8}{*}{\cellcolor[HTML]{DED06E}{\color[HTML]{FFFFFF} \begin{tabular}[c]{@{}c@{}}1\\ (Sec.6)\end{tabular}}} &  \multirow{-4}{*}{\cellcolor[HTML]{DDD06E}{\color[HTML]{FFFFFF} \textbf{\begin{tabular}[c]{@{}c@{}}Augmentation\\ (Sec. 6.2)\end{tabular}}}} &  \cellcolor[HTML]{FFF2CC}Depth Estimation &  Comic Panels/Images &  Depth Map \\

\midrule

\cellcolor[HTML]{E8A84D}{\color[HTML]{FFFFFF} } &  \cellcolor[HTML]{FFD966}{\color[HTML]{FFFFFF} } &  \cellcolor[HTML]{FFE599}Object Detection &  \cellcolor[HTML]{EFEFEF}Tag/s + Image &  \cellcolor[HTML]{EFEFEF}Bounding Boxes \\

\cellcolor[HTML]{E8A84D}{\color[HTML]{FFFFFF} } &  \cellcolor[HTML]{FFD966}{\color[HTML]{FFFFFF} } &  \cellcolor[HTML]{FFE599}Character Re-identification (Character ID) &  Multiple Panels &  Tag \\

\cellcolor[HTML]{E8A84D}{\color[HTML]{FFFFFF} } &  \multirow{-3}{*}{\cellcolor[HTML]{FFD966}{\color[HTML]{FFFFFF} \textbf{\begin{tabular}[c]{@{}c@{}}Grounding\\ (Sec. 7.1)\end{tabular}}}} &  \cellcolor[HTML]{FFE599}Grounding (IG) &  \cellcolor[HTML]{EFEFEF}{\color[HTML]{000000} {[}Prompt{]} + Image} &  \cellcolor[HTML]{EFEFEF}Bounding Boxes \\

\cellcolor[HTML]{E8A84D}{\color[HTML]{FFFFFF} } &  \cellcolor[HTML]{F6B26B}{\color[HTML]{FFFFFF} } &  \cellcolor[HTML]{F9CB9C}Character-Balloon Association (Speaker ID) &  Character + Balloons &  Tag \\

\cellcolor[HTML]{E8A84D}{\color[HTML]{FFFFFF} } &  \cellcolor[HTML]{F6B26B}{\color[HTML]{FFFFFF} } &  \cellcolor[HTML]{F9CB9C}Dialog transcription &  \cellcolor[HTML]{EFEFEF}Image &  \cellcolor[HTML]{EFEFEF}Text \\

\cellcolor[HTML]{E8A84D}{\color[HTML]{FFFFFF} } &   \multirow{-3}{*}{\cellcolor[HTML]{F6B26B}{\color[HTML]{FFFFFF} \textbf{\begin{tabular}[c]{@{}c@{}}Analysis\\ (Sec. 7.2)\end{tabular}}}}  &  \cellcolor[HTML]{F9CB9C}Translation &  Image &  Text \\

\multirow{-8}{*}{\cellcolor[HTML]{E8A84D}{\color[HTML]{FFFFFF} \begin{tabular}[c]{@{}c@{}}2\\ (Sec.7)\end{tabular}}} &  \cellcolor[HTML]{E69138}{\color[HTML]{FFFFFF} \textbf{\begin{tabular}[c]{@{}c@{}}Segmentation\\ (Sec. 7.3)\end{tabular}}} &  \cellcolor[HTML]{F6B26B}Instance Segmentation (IS) &  \cellcolor[HTML]{EFEFEF}{\color[HTML]{000000} {[}Prompt{]} + Image} &  \cellcolor[HTML]{EFEFEF}Segments \\

\midrule






\cellcolor[HTML]{E06666}{\color[HTML]{FFFFFF} } &  \cellcolor[HTML]{E06666}{\color[HTML]{FFFFFF} } &  \cellcolor[HTML]{EAD1DC}Image-Text Retrieval (IR) &  Text &  Image \\

\cellcolor[HTML]{E06666}{\color[HTML]{FFFFFF} } &  \cellcolor[HTML]{E06666}{\color[HTML]{FFFFFF} } &  \cellcolor[HTML]{EAD1DC}Text-Image Retrieval (TR) &  \cellcolor[HTML]{EFEFEF}Image &  \cellcolor[HTML]{EFEFEF}Text \\

\cellcolor[HTML]{E06666}{\color[HTML]{FFFFFF} } &  \multirow{-3}{*}{\cellcolor[HTML]{E06666}{\color[HTML]{FFFFFF} \textbf{\begin{tabular}[c]{@{}c@{}}Retrieval\\ (Sec. 8.1)\end{tabular}}}} &   \cellcolor[HTML]{EAD1DC}Composed Image Retrieval (CIR) &  Text + Image &  Image \\

\cellcolor[HTML]{E06666}{\color[HTML]{FFFFFF} } &  \cellcolor[HTML]{CC4125}{\color[HTML]{FFFFFF}} & \cellcolor[HTML]{EA9999}Image Inpainting (II) &  \cellcolor[HTML]{EFEFEF}{\color[HTML]{000000} Text + {[}prompt{]} + Image} &  \cellcolor[HTML]{EFEFEF}Image \\

\multirow{-5}{*}{\cellcolor[HTML]{E06666}{\color[HTML]{FFFFFF} \begin{tabular}[c]{@{}c@{}}3\\ (Sec.8)\end{tabular}}} &  \multirow{-2}{*}{\cellcolor[HTML]{CC4125}{\color[HTML]{FFFFFF} \textbf{\begin{tabular}[c]{@{}c@{}}Modification\\ (Sec. 8.2)\end{tabular}}}} &  \cellcolor[HTML]{EA9999}Image Editing via Text (IET) &  Text + Image &  Image \\

\midrule


\cellcolor[HTML]{CC4125}{\color[HTML]{FFFFFF} } &  \cellcolor[HTML]{990000}{\color[HTML]{FFFFFF} } &  \cellcolor[HTML]{E06666}Visual Question Answering (VQA) &  Image + Text &  Text \\

\cellcolor[HTML]{CC4125}{\color[HTML]{FFFFFF} } &  \cellcolor[HTML]{990000}{\color[HTML]{FFFFFF} } &  \cellcolor[HTML]{E06666}Visual Dialog (VisDial) &  \cellcolor[HTML]{EFEFEF}Image + Dialog + Text &  \cellcolor[HTML]{EFEFEF}Text \\

\multirow{-3}{*}{\cellcolor[HTML]{CC4125}{\color[HTML]{FFFFFF} \begin{tabular}[c]{@{}c@{}}4\\ (Sec.9)\end{tabular}}}  &  \multirow{-3}{*}{\cellcolor[HTML]{990000}{\color[HTML]{FFFFFF} \textbf{\begin{tabular}[c]{@{}c@{}}Understanding\\ (Sec. 9.1)\end{tabular}}}} &  \cellcolor[HTML]{E06666}Visual Reasoning (VR) &  Image + Text &  Text \\

\midrule

\cellcolor[HTML]{A61C00}{\color[HTML]{FFFFFF} } &  \cellcolor[HTML]{A61C00}{\color[HTML]{FFFFFF} } &  \cellcolor[HTML]{DD7E6B}Image-2-Text Generation (I2T) &  \cellcolor[HTML]{EFEFEF}Image &  \cellcolor[HTML]{EFEFEF}Text \\

\cellcolor[HTML]{A61C00}{\color[HTML]{FFFFFF} } &  \cellcolor[HTML]{A61C00}{\color[HTML]{FFFFFF} } &  \cellcolor[HTML]{DD7E6B}Grounded Image Captioning (GIC) &  Image &  Text + Bbox \\

\cellcolor[HTML]{A61C00}{\color[HTML]{FFFFFF} } &  \cellcolor[HTML]{A61C00}{\color[HTML]{FFFFFF} } &  \cellcolor[HTML]{DD7E6B}Text-2-Image Generation (T2I) &  \cellcolor[HTML]{EFEFEF}Text &  \cellcolor[HTML]{EFEFEF}Image \\

\cellcolor[HTML]{A61C00}{\color[HTML]{FFFFFF} } & \cellcolor[HTML]{A61C00}{\color[HTML]{FFFFFF} }  &  \cellcolor[HTML]{DD7E6B}Scene Graph Generation for Captioning &  Comic Panel &  Scene Graph \\

\cellcolor[HTML]{A61C00}{\color[HTML]{FFFFFF} } &  \multirow{-5}{*}{\cellcolor[HTML]{A61C00}{\color[HTML]{FFFFFF} \textbf{\begin{tabular}[c]{@{}c@{}}Generation\\ (Sec. 10.1)\end{tabular}}}} &  \cellcolor[HTML]{DD7E6B}Sound Generation from Single Panel &  \cellcolor[HTML]{EFEFEF}Single Comic Panel &  \cellcolor[HTML]{EFEFEF}Sound/Audio \\

\cellcolor[HTML]{A61C00}{\color[HTML]{FFFFFF} } & \cellcolor[HTML]{741B47}{\color[HTML]{FFFFFF} }  &  \cellcolor[HTML]{C27BA0}3D Model Generation from Images (3DGI) &  Collection of Images &  3D Model \\

\cellcolor[HTML]{A61C00}{\color[HTML]{FFFFFF} } &  \cellcolor[HTML]{741B47}{\color[HTML]{FFFFFF} } &  \cellcolor[HTML]{C27BA0}Video Generation from Text (VGT) &  \cellcolor[HTML]{EFEFEF}Complex Long Text &
  \cellcolor[HTML]{EFEFEF}Video \\

\multirow{-8}{*}{\cellcolor[HTML]{A61C00}{\color[HTML]{FFFFFF} \begin{tabular}[c]{@{}c@{}}5\\ (Sec.10)\end{tabular}}} & \multirow{-3}{*}{\cellcolor[HTML]{741B47}{\color[HTML]{FFFFFF} \textbf{\begin{tabular}[c]{@{}c@{}}Synthesis\\ (Sec. 10.2)\end{tabular}}}} &  \cellcolor[HTML]{C27BA0}Narrative-Based Complex Scene Generation (NCSG) &   Detailed Narrative Text &
  Series of Images \\
\bottomrule
\end{tabular}
\label{tab:locu}
\end{table*}
%


\subsection{Relevant Surveys}

To the best of our knowledge, this is the \textit{first} survey that reviews comics understanding publications with a task-oriented approach for Vision-Language. Several relevant surveys have been conducted, examining broader aspects of visual art \cite{santos_artificial_2021} and, more specifically, comics \cite{augereau_overview_2017, augereau_survey_2018, nairat_generative_2021}. Notably, Augereau et al. \cite{augereau_survey_2018} categorized comic research into three primary areas: (i) content analysis, (ii) content generation and adaptation, and (iii) user interaction, which remain relevant entry points for understanding this field's research and applications.

In contrast, building on recent advancements, our survey focuses on two key aspects: 1) the recent progress in \comicsunderstanding~tasks and datasets, and 2) the application of Vision-Language tasks to the unique characteristics of comics. 

\section{Comics Foundations}\label{sec:foundation}

Our preliminary analysis prioritizes understanding the structural intricacies of comics.
The term ``comics'' broadly denotes the entire medium in its uncountable form, while ``a comic'' refers to a singular entity within this medium, such as a comic book or strip. 
The structure of comic books has been a subject of extensive study, examining aspects like their commonalities, layout structures \cite{cohn_visual_2023}, content arrangement, narrative transitions \cite{cohn_speech_2013, cohn_visual_2013, cohn_cognition_2014, cohn_architecture_2014, cohn_multimodal_2016}, and stylistic narrative variations \cite{cohn_picture_2017, cohn_who_2020, klomberg_running_2022}. 

\begin{figure*}
    \centering
    \includegraphics[width=\linewidth]{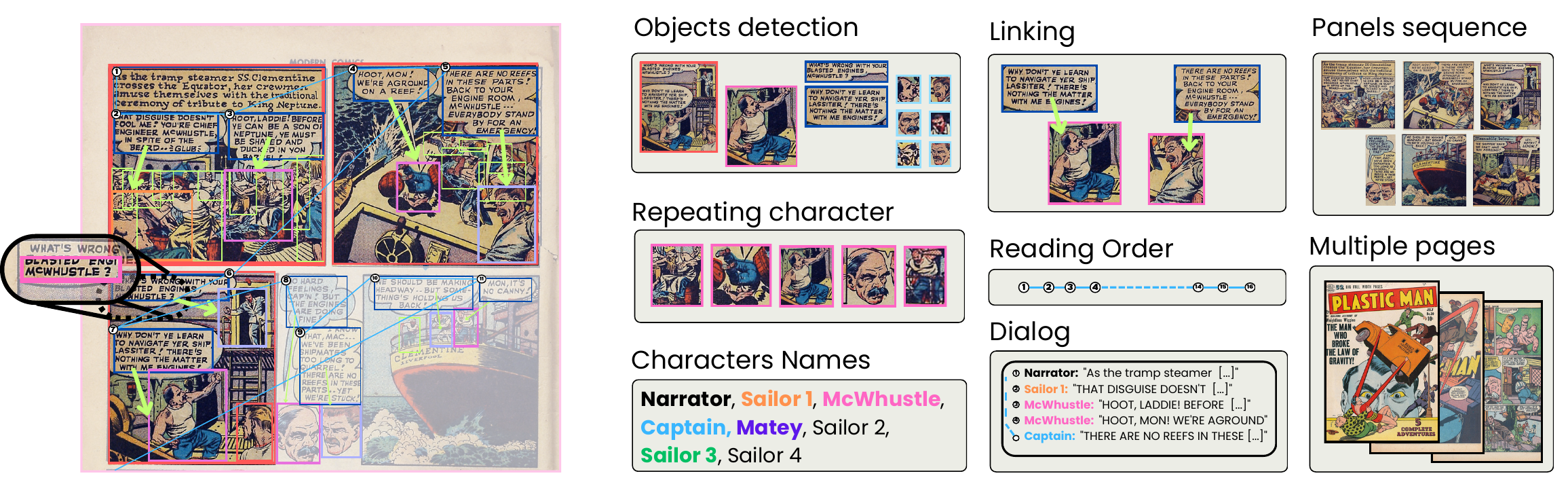}
    \caption{Anatomy of Comic Page Elements: This illustration delineates the various components found on comic pages including object detection, linkages between elements, the designated reading order of texts and panels, as well as key textual features like character names—all arranged in a logical sequence throughout the panels and pages.}
    \label{fig:comics-elements}
\end{figure*}

Typically, comics are presented in single or double-page formats, influenced by the artist’s style. They narrate stories across various genres or series, such as superheroes and science fiction tales, sometimes interlinking through crossovers. Notably, artistic consistency within a series or genre, often achieved through collaborative efforts, is a key feature. Each comic page (as illustrated in Figure \ref{fig:comics-elements}) encapsulates individual scenes within panels of diverse types - from splash panels for opening scenes to widescreen panels for expansive narratives. These panels convey interactions and dialogues through speech or thought balloons, and expressive onomatopoeia, enriching the narrative experience.
Artistic variations in style and color also play a significant role in comics, ranging from detailed realism to abstract forms, with the color palette setting the mood and atmosphere.
Despite rigorous analyses \cite{cohn_different_2011}, no significant differences have been identified in terms of attention, subjectivity, and viewpoint between Manga and American comics. Hence, these distinctions 
are better explored within their distinct historical, cultural, and artistic contexts:

\textbf{American Comics}: With a rich history dating back to the 1930s, American comics are celebrated for their colorful and high-quality presentations, primarily known for their superhero narratives. They adhere to a top-to-bottom, left-to-right reading format and are characterized by a dialogue-rich and brisk storytelling style.

\textbf{Manga}: Emerging in post-World War II Japan and influenced by diverse sources, including Japanese traditional art and American comics, manga is recognized for its unique black-and-white style, though colored variants exist. Covering a broad range of genres, Manga is read right-to-left, top-to-bottom. This style is marked by a distinctive ``Japanese Visual Language'' that is consistent across genres \cite{cohn_vocabulary_2016}.

%
%

\section{Tasks and Datasets}\label{sec:dataset}

This section summarizes the commonly used Comics datasets, as detailed in Table~\ref{tab:datasets}. Some of these datasets are either intended for training or evaluation, depending on their sizes.

\renewcommand\arraystretch{0.95}
\begin{table*}[t]
\tiny
\centering
    \caption{Overview of Comic/Manga Datasets and Tasks, including information on availability, published year, source, and properties (languages, number of books and pages). Accessibility is marked with {\color{red} \xmark} for no longer existing datasets, {\color{gray} \xmark} for existing but inaccessible datasets, and {\color{ForestGreen} \cmark} for accessible ones. The link \textcolor{magenta}{\faGlobe} directs to project websites, while \textcolor{blue}{\faDatabase} links to dataset websites. The tasks identifiers corresponds to Image Captioning (IC), Object Detection (OD), Re-Identification (RI), Linking (L), Segmentation (S), Dialog Generation (DG), and Analysis (A).}
    \begin{tabular}{l|l|c|c|c|c|c|c|c@{\hskip 0.1in}c@{\hskip 0.1in}c@{\hskip 0.1in}c@{\hskip 0.1in}c@{\hskip 0.1in}c@{\hskip 0.1in}c}
    \toprule
    \textbf{From} & \textbf{Name} & \textbf{Year} & \textbf{Acc.} & \textbf{Lang.} & \textbf{Origin} & \textbf{\# books} & \textbf{\# pages} & \textbf{IC} & \textbf{OD} & \textbf{RI} & \textbf{L} & \textbf{S} & \textbf{DG} & \textbf{A} \\ 
    \midrule
    \rowcolors{2}{lightgray}{}

    & \textbf{Fahad18 \cite{khan_color_2012}} \href{http://www.cat.uab.cat/Research/object-detection/}{\faGlobe} & 2012 & {\color{gray} \xmark} & - & - & - & 586 &  & x & x &  &  &  &  \\ 
    
    \rowcolor[HTML]{F3F3F3}
    & \textbf{eBDtheque \cite{guerin_ebdtheque_2013}} \href{https://ebdtheque.univ-lr.fr/}{\faGlobe} \href{https://ebdtheque.univ-lr.fr/registration/}{\textcolor{blue}{\faDatabase}} & 2013 & {\color{ForestGreen} \cmark} & GB, FR, JP & 1905-2012 & 25 & 100 &  & x &  & x &  &  &  \\
    \cite{guerin_ebdtheque_2013} & \textbf{sun70 \cite{sun_specific_2013}} & 2013 & {\color{gray} \xmark} & FR & - & 6 & 60 &  & x &  & x &  &  &  \\
    \rowcolor[HTML]{F3F3F3} 
    & \textbf{Ho42 \cite{ho_redundant_2013_new}} & 2013 & {\color{gray} \xmark} & - & - & - & 42 &  &  & x &  &  &  &  \\ 
    
    \cite{guerin_ebdtheque_2013} & \textbf{SSGCI \cite{le_subgraph_2018}} \href{http://icpr2016-ssgci.univ-lr.fr/challenge/dataset-download/}{\faGlobe} \href{http://icpr2016-ssgci.univ-lr.fr/challenge/dataset-download/}{\textcolor{blue}{\faDatabase}} & 2016 & {\color{gray} \xmark} & GB, FR, JP & 1905-2012 & - & 500 &  & x &  &  &  &  &  \\ 
    
    \rowcolor[HTML]{F3F3F3} 
    & \textbf{Sequencity \cite{nguyen_comic_2017}} & 2017 & {\color{red} \xmark} & GB, JP & - & - & 140000 & x &  &  &  &  &  &  \\
    & \textbf{BAM! \cite{wilber_bam_2017}} & 2017 & {\color{red} \xmark} & - & - & - & 2500000 & x & x &  &  &  &  &  \\
    \rowcolor[HTML]{F3F3F3} 
    & \textbf{COMICS \cite{iyyer_amazing_2017}} \href{https://github.com/miyyer/comics}{\faGlobe} \href{https://obj.umiacs.umd.edu/comics/index.html}{\textcolor{blue}{\faDatabase}} & 2017 & {\color{ForestGreen} \cmark} & GB & 1938-1954 & 3948 & 198657 &  & x &  &  &  &  &  \\
    \cite{fujimoto_manga109_2016} & \textbf{JC2463 \cite{qin_faster_2017}} & 2017 & {\color{gray} \xmark} & JP & - & 14 & 2463 &  & x &  &  &  &  &  \\
    \rowcolor[HTML]{F3F3F3} 
    & \textbf{AEC912 \cite{qin_faster_2017}} & 2017 & {\color{gray} \xmark} & GB, FR & - & - & 912 &  & x &  &  &  &  &  \\
    & \textbf{GCN \cite{dunst_graphic_2017}} \href{https://groups.uni-paderborn.de/graphic-literature/gncorpus/corpus.php}{\faGlobe} \href{https://groups.uni-paderborn.de/graphic-literature/gncorpus/download.php}{\textcolor{blue}{\faDatabase}} & 2017 & {\color{gray} \xmark} & GB, JP & 1978-2013 & 253 & 38000 &  & x &  & x &  &  &  \\
    \rowcolor[HTML]{F3F3F3} 
    \cite{nguyen_comic_2017} & \textbf{Sequencity612 \cite{nguyen_comic_2017}} & 2017 & {\color{red} \xmark} & GB, JP & - & - & 612 &  & x &  &  &  &  &  \\
    \cite{fujimoto_manga109_2016} & \textbf{Comics3w \cite{he_endtoend_2018}} \href{https://philokey.github.io/sren.html}{\faGlobe} & 2017 & {\color{gray} \xmark} & JP, GB & - & 103 & 29845 &  & x &  &  &  &  &  \\ 
    
    \rowcolor[HTML]{F3F3F3} 
    & \textbf{comics2k \cite{inoue_crossdomain_2018}} \href{https://naoto0804.github.io/cross_domain_detection/}{\faGlobe} \href{https://github.com/naoto0804/cross-domain-detection/tree/master/datasets}{\textcolor{blue}{\faDatabase}} & 2018 & {\color{red} \xmark} & - & - & - & - &  & x &  &  &  &  &  \\
    & \textbf{DCM772 \cite{nguyen_digital_2018}} \href{https://paperswithcode.com/dataset/dcm}{\faGlobe} \href{https://git.univ-lr.fr/crigau02/dcm_dataset}{\textcolor{blue}{\faDatabase}} & 2018 & {\color{ForestGreen} \cmark} & GB & 1938-1954 & 27 & 772 &  & x &  &  &  &  &  \\
    \rowcolor[HTML]{F3F3F3} 
    \cite{fujimoto_manga109_2016} & \textbf{Manga109 \cite{fujimoto_manga109_2016,ogawa_object_2018}} \href{http://www.manga109.org/en/download.html}{\faGlobe} \href{http://www.manga109.org/en/download.html}{\textcolor{blue}{\faDatabase}} & 2018 & {\color{ForestGreen} \cmark} & JP & 1970-2010 & 109 & 21142 & x & x & x & x &  &  &  \\ 
    
    \cite{nguyen_comic_2017} & \textbf{Sequencity4k \cite{nguyen_learning_2020}} & 2020 & {\color{red} \xmark} & GB, FR, JP & - & - & 4479 &  &  &  &  & x &  &  \\ 
    
    \rowcolor[HTML]{F3F3F3} 
    \cite{iyyer_amazing_2017} & \textbf{EmoRecCom \cite{nguyen_icdar_2021}} \href{https://sites.google.com/view/emotion-recognition-for-comics}{\faGlobe} \href{https://competitions.codalab.org/competitions/30954\#participate-get_data}{\textcolor{blue}{\faDatabase}} & 2021 & {\color{ForestGreen} \cmark} & GB & 1938-1954 & - & - & x &  &  &  &  &  &  \\ 
    
    & \textbf{BCBId \cite{dutta_bcbid_2022}} \href{https://sites.google.com/view/banglacomicbookdataset}{\faGlobe} \href{https://sites.google.com/view/banglacomicbookdataset/contacts?authuser=0}{\textcolor{blue}{\faDatabase}} & 2022 & {\color{ForestGreen} \cmark} & BD & - & 64 & 3327 &  & x &  &  &  &  &  \\
    \rowcolor[HTML]{F3F3F3} 
    \cite{fujimoto_manga109_2016} & \textbf{COO \cite{baek_coo_2022}} \href{https://github.com/ku21fan/COO-Comic-Onomatopoeia}{\faGlobe} \href{https://github.com/manga109/public-annotations\#comic-onomatxopoeia-coo}{\textcolor{blue}{\faDatabase}} & 2022 & {\color{ForestGreen} \cmark} & JP & 1970-2010 & 109 & 10602 &  & x &  &  &  &  &  \\
    \cite{iyyer_amazing_2017} & \textbf{COMICS-Text+ \cite{soykan_comprehensive_2022}} \href{https://github.com/gsoykan/comics_text_plus}{\faGlobe} \href{https://github.com/gsoykan/comics_text_plus\#getting-started}{\textcolor{blue}{\faDatabase}} & 2022 & {\color{ForestGreen} \cmark} & GB & 1938-1954 & 3948 & 198657 &  & x &  &  &  &  &  \\ 
    
    \rowcolor[HTML]{F3F3F3} 
    & \textbf{VLRC \cite{cohn_visual_2023}} \href{https://dataverse.nl/}{\faGlobe} \href{https://dataverse.nl/dataset.xhtml?persistentId=doi:10.34894/LWMZ7G}{\textcolor{blue}{\faDatabase}} & 2023 & {\color{gray} \xmark} & JP, FR, GB, 6+ & 1940-present & 376 & 7773 &  &  &  &  &  &  & x \\ 
    
    & \textbf{PopManga \cite{sachdeva_manga_2024}} \href{https://github.com/gsoykan/comics_text_plus}{\faGlobe} \href{https://github.com/gsoykan/comics_text_plus\#getting-started}{\textcolor{blue}{\faDatabase}} & 2024 & {\color{ForestGreen} \cmark} & GB & 1990-2020 & 25 & 1925 &  & x & x & x &  & x &  \\
    \rowcolor[HTML]{F3F3F3} 
    mix & \textbf{CoMix \cite{vivoli_comix_2024}} \href{https://emanuelevivoli.github.io/CoMix-dataset}{\faGlobe} \href{https://rrc.cvc.uab.es/?ch=31}{\textcolor{blue}{\faDatabase}} & 2024 & {\color{ForestGreen} \cmark} & GB, FR & 1938-2023 & 100 & 3800 &  & x & x & x &  & x &  \\

    \bottomrule[1pt]
    \end{tabular}
    \label{tab:datasets}
\end{table*}

\subsection{Annotations overview}
The available current comics datasets were introduced for different tasks. From Table~\ref{tab:datasets} it is noticeable that the available datasets vary significantly in size, languages, origin, and -more importantly- annotations, making them cover different tasks.

eBDtheque \cite{guerin_ebdtheque_2013} offers comprehensive annotations in high quality for Character, Balloon, and Panel detection, along with Text-Character association. This makes it usable for object detection and linking tasks but it comes with a drawback: it is composed of only 100 pages. In contrast, the COMICS dataset \cite{iyyer_amazing_2017} focuses on a set of tasks called closure, based on automatic panel and text detection over a large number of pages. However, it was introduced with poor annotations. Only recently some works \cite{soykan_comprehensive_2022} have improved OCR text detection on the COMICS dataset. In the same vein, \cite{vivoli_multimodal_2024} shows that improving OCR accuracy (using Textract) could lead to marginal accuracy gain in closure tasks for the COMICS dataset. Both \cite{soykan_comprehensive_2022,vivoli_multimodal_2024} released the novel OCR annotations to the research community.
It is to note also that some of the dataset with high-quality annotations and large number of pages are not accessible. For instance,  BCBId \cite{dutta_bcbid_2022}, though promising high-quality annotations, falls short in delivery.  Similarly, VLRC \cite{cohn_visual_2023}, while rich in annotations, lacks digital applicability due to its focus on physical books. 
Manga109~\cite{fujimoto_manga109_2016} is the first to provide high-level annotations, covering more than 20k pages. The dataset has been improved over several iterations, further expanding the scope of the dataset by adding annotations for onomatopoeias \cite{baek_coo_2022} and dialog \cite{li_manga109dialog_2023}. The current state of Manga109 is the result of many years of annotation and improvement by subsequent developments. More recently, following the Manga109 wave, PopManga~\cite{sachdeva_manga_2024} has been proposed as an English-language alternative, including annotations for recognition and linking, but on a smaller scale (2k images).

In summary, while datasets for \comicsunderstanding~exist, there is a need for more approaches to standardize and unify their annotations. 
The Comics Dataset Framework~\cite{vivoli_cdf_2024} proposed a unification of detection annotations as well as a standardization of metrics and settings used to train and measure models. The work was subsequently extended in CoMix~\cite{vivoli_comix_2024}, where mainly English-language comics (and manga) were selected and the annotations extended to the task of single-page dialogue transcription. In this work, performances of several state-of-the-art models, including Magi~\cite{sachdeva_manga_2024}, CLIP~\cite{radford_learning_2021}, DINOv2~\cite{oquab_dinov2_2023}, and GPT4~\cite{openai_gpt4_2024}, were measured, across various datasets, for the first time.

However, as highlighted in the first column of Table~\ref{tab:datasets}, the complexity of annotations and tasks is limited. While object detection and linking are integral tasks to understanding the narrative and dialogue in comics, we must advance toward more complex tasks found in Vision-Language research. These will be discussed within our taxonomy in the next section.

\subsection{Datasets overview}

The domain of \comicsunderstanding~encompasses a range of datasets focused on processing and examining comic media. A review of various datasets, as detailed in Table \ref{tab:datasets}, reveals that several are no longer accessible (marked with {\color{red} \xmark}) or have restricted access ({\color{gray} \xmark}). Those available ({\color{ForestGreen} \cmark}) often require 
\begin{wrapfigure}{rb}{0.6\textwidth}
\vspace{0.1in}
    \centering
    \includegraphics[width=0.6\textwidth]{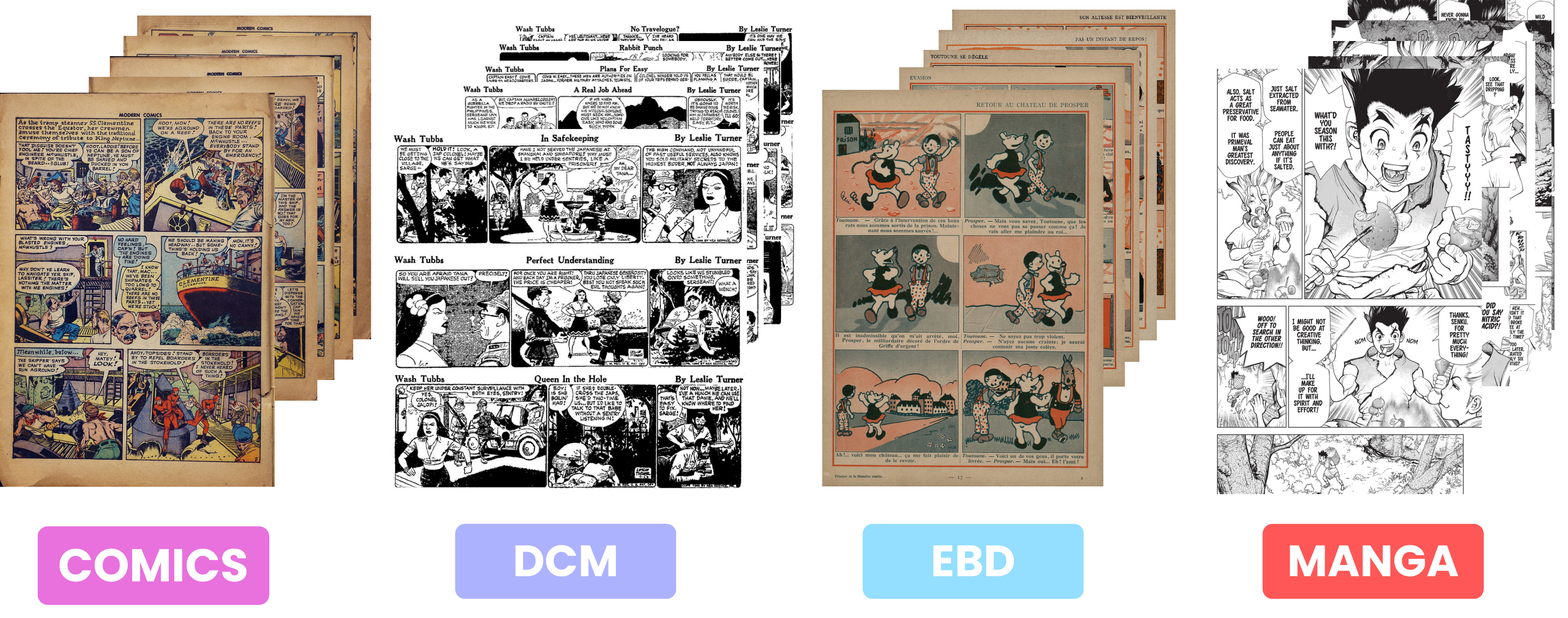}
    \caption{ Example of different comic datasets, from Black-and-White to Color, from comics-style to Manga-style.}
    \label{fig:styles}
\vspace{-0.1in}
\end{wrapfigure}
approval from respective research groups to obtain them.
This is a common practice in the field of ``sharing copyrighted materials'' as in fact they can be used only for research purposes in the majority of cases. If the copyright does not allow for redistribution, it is a common practice to release annotations and a script to download and structure the dataset accordingly (as for \cite{sachdeva_manga_2024}).
As we can see from Table \ref{tab:datasets}, many datasets share common origins (column \textit{``From''}).
Notably, these datasets vary significantly in terms of style, quality, and size. The common languages of these datasets are English, French, and Japanese and none of the available data seems to be collected from recent books (after 2010), probably due to copyright limits. We found the above limitation to be one of the main issues in the comic domain: (almost) every work proposes its own (sometimes private) dataset on which results are drawn with little comparison, making these works hard to replicate. Datasets are limited in size and the model's \textit{code and weights} are rarely shared. Luckily, this has started changing recently \cite{vivoli_cdf_2024,sachdeva_manga_2024,vivoli_comix_2024}. 

\begin{wraptable}{r}{0.55\textwidth} 
\tiny
\centering
\caption{Number of single instance annotation (declared) in papers. The ``Obj. cls'' stands for Object-based classification, and refers to the number of classes available to classify the detected instances in fine-grained classes. Colored images are indicated with \artColor, while black and white images with \artBW. The only dataset that provide Onomatopoeias annotations is PopManga (61465).}
\begin{tabular}{l|c|c|c|c|c}
\toprule

\textbf{Mode / Name} & \textbf{Panel} & \textbf{Char.} & \textbf{Face} & \textbf{Text} & \textbf{Cls} \\

\midrule

\artColor~\textbf{Fahad18 \cite{khan_color_2012}} & - & 18 & - & - & 18 \\
\rowcolor[HTML]{F3F3F3} 
\artColor~\textbf{Ho42 \cite{ho_redundant_2013_new}} & 200 & 100 & - & - & - \\

\artColor~\textbf{eBDtheque \cite{guerin_ebdtheque_2013}} & 850 & 1550 & - & 1092 & 5 \\

\rowcolor[HTML]{F3F3F3} 
\artColor~\textbf{SSGCI \cite{le_subgraph_2018}} & - & 50 & - & - & 1  \\

\artColor~\textbf{COMICS \cite{iyyer_amazing_2017,soykan_comprehensive_2022}} & 1229664 & - & - & 2498657 & 2  \\

\rowcolor[HTML]{F3F3F3} 
\artBW~\textbf{Comics3w \cite{he_sren_2017}} & 159529 & - & - & - & 1 \\

\artBW~\textbf{JC2463 \cite{qin_faster_2017}} & - & - & 15801 & - & 2 \\

\rowcolor[HTML]{F3F3F3} 
\artColor~\textbf{AEC912 \cite{qin_faster_2017}} & - & - & 8184 & - & 2 \\

\artColor~\textbf{DCM772 \cite{nguyen_digital_2018}} & 4470 & 10757 & 5438 & - & 4  \\

\rowcolor[HTML]{F3F3F3} 
\artBW~\textbf{Manga109 \cite{fujimoto_manga109_2016,ogawa_object_2018,baek_coo_2022}} & 103900 & 157152 & 118715 & 147918 & 4 \\

\artColor~\textbf{EmoRecCom \cite{nguyen_icdar_2021}} & 10199 & - & - & - & 8  \\

\rowcolor[HTML]{F3F3F3}
\artBW~\textbf{PopManga \cite{sachdeva_manga_2024}} & - & 18783 & - & 20843 & 2 \\ 

\artBW/\artColor~\textbf{CoMix \cite{vivoli_comix_2024}} & 22176 & 48903 & 32625 & 37923 & 4 \\ 
\bottomrule
\end{tabular}
\label{tab:db-details}
\vspace{-0.1in}
\end{wraptable}

In Table \ref{tab:db-details} we report, for every dataset, the declared numbers of annotations available. On the right-most column, we report the number of classes annotated, which correspond to different class objects for the detection, or the number of classes for classification annotations. For instance, in \textit{Fahad18} there are 18 characters, thus the number of classes is 18. However, in \textit{eBDtheque}, there are 5 classes, but only panel, character, and face annotations. This is because the characters are separated into three sub-categories: human-like, animal-like, and object-like.  Here, we describe the datasets chronologically and provide an overview of the image style in Figure \ref{fig:styles}.

\textbf{eBDtheque:} The eBDtheque\footnote{\href{http://ebdtheque.univ-lr.fr/registration}{http://ebdtheque.univ-lr.fr/registration}} \cite{guerin_ebdtheque_2013} includes 100 pages from 20 books, mainly in French, with some English and Japanese comics. \cite{rigaud_segmentation_2016} labeled 850 panels, 1,092 balloons, 1,550 characters, and 4,691 text lines, offering expert annotations. Balloon labels include instance segmentation and classification (speech, thought, narration, illustrative).

\textbf{Manga109:} The Manga109\footnote{\href{http://www.manga109.org/index_en.php}{http://www.manga109.org/index\_en.php}} \cite{fujimoto_manga109_2016} dataset includes 109 manga volumes from 93 authors, with copyrighted material obtained under a ``restrictive term-of-use'', allowing use for research and publications with proper citation. The mangas, published between the 1970s and 2010s, span 12 genres. The dataset was extended by COO \cite{baek_coo_2022}, which added onomatopoeia annotations (5.8 per page on average), and Manga-Dialog \cite{li_manga109dialog_2023}, which linked text boxes to characters.

\textbf{COMICS:} The COMICS\footnote{\href{https://obj.umiacs.umd.edu/comics/index.html}{https://obj.umiacs.umd.edu/comics/index.html}} \cite{iyyer_amazing_2017} dataset contains 1,229,664 panels with automatic textbox transcriptions from golden-age American comics (1938-1954). It includes 3,948 books from the Digital Comics Museum and manually annotated bounding boxes for 500 pages and 1,500 textboxes. Later works improved OCR transcriptions \cite{soykan_comprehensive_2022,vivoli_multimodal_2024}, with pipelines developed for better OCR processing and transcript extraction. \cite{agrawal_multimodal_2023} added dialog prediction based on "persona" information.

\textbf{DCM772:} The DCM772\footnote{\href{https://git.univ-lr.fr/crigau02/dcm_dataset}{https://git.univ-lr.fr/crigau02/dcm\_dataset}} \cite{nguyen_digital_2018}  contains 772 (pages) images from 27 golden-age comic books from the same COMICS source. Annotations, however, are more precise as they contain panels, the associated characters bounding boxes, and also face boxes.

\subsection{Datasets for Evaluation}

Across the datasets presented in Table~\ref{tab:datasets}, two recent datasets are provided as evaluation sets. In particular, PopManga proposes two splits (validation and test) while CoMix enhances annotations for various existing datasets among which also PopManga.

\textbf{PopManga:} The PopManga\footnote{\href{https://github.com/ragavsachdeva/Magi/tree/main/datasets}{https://github.com/ragavsachdeva/Magi}} \cite{sachdeva_manga_2024} dataset contains English manga titles from the most popular mangas. The dataset contains two test splits: test-seen with 1,136 pages and test-unseen with 789 pages, obtained from 15 and 10 books respectively. The test sets have been annotated with Text (both textboxes and onomatopoeias annotated as text), and Characters.

\textbf{CoMix:} The CoMix\footnote{\href{https://emanuelevivoli.github.io/CoMix-dataset}{https://emanuelevivoli.github.io/CoMix-dataset}} \cite{vivoli_comix_2024} dataset contains English and French comics and manga titles from the most popular datasets: eBDtheque, COMICS, DCM, and PopManga. The dataset contains two splits: validation (available with annotations) and the held-out test split (available only through the evaluation server\footnote{Accessible at \href{https://rrc.cvc.uab.es/?ch=31}{Robust Reading Competition} website}). It contains 3.8k images across 100 books, and annotations for multi-task, from detection and linking to dialog generation.

\section{Taxonomy: \textit{LoCU}}\label{sec:locu}

Developing a comprehensive taxonomy for comics within the Vision-Language (VL) domain requires a careful overview of existing work in VL surveys and tasks. Our approach synthesizes insights from various benchmarks and taxonomies, forming the foundation for our unique classification – the \textit{Layers of \comicsunderstanding} (\textit{LoCU}).

\textbf{Relevant Taxonomies.} Inspiration for this taxonomy comes from multimodal benchmarks like MMMU \cite{yue_mmmu_2023}, which highlight the need for a structure addressing modal complexity and task-specific intricacies in comics. The dimensional analysis in video understanding from \cite{fan_aigcbench_2024} also informs our spatial and temporal considerations for comics.
In VL evaluation, \cite{fu_mme_2024} reframes tasks as Visual Question Answering (VQA), but lacks a comprehensive structure for comics-related tasks. Challenges in multimodal learning from \cite{baltrusaitis_multimodal_2017} guide our classification of VL tasks, with a task-centric focus.
\cite{li_visionlanguage_2022} provides a foundational classification of VL tasks into grounding, retrieval, understanding, and generation. Their work is crucial in shaping our approach, and we build upon their framework by extending these categories to account for comic-specific tasks. In particular, we add categories like tagging, augmentation, analysis, segmentation, modification, and synthesis, which better capture the diversity of tasks observed in comic studies and expand the structure to fully represent the unique challenges in comics understanding.

In Table \ref{tab:locu}, we present 31 Vision-Language multimodal tasks, categorized according to the modalities involved, forming the backbone of the \textit{LoCU} framework. This taxonomy, spanning 10 distinct task groups across five layers, aims to capture the multifaceted nature of \comicsunderstanding. Layer 0, representing fundamental image processing and viewing capabilities, sets the stage for increasingly complex interactions in the subsequent layers. More details are provided in Supplementary Materials. 
This structured approach allows for a comprehensive and nuanced understanding of the range and complexity of tasks within the comics domain, laying a foundation for future exploration and innovation in comic analysis and understanding.

\section{Layer 1: Tagging and Augmentation}
\label{sec:locu-layer1}

The first layer comprises tasks with unimodal input (single image or sequence of images) and unimodal output (Images or Text): \textit{Tagging} and \textit{Augmentation}.

\subsection{Tagging}
\begin{wrapfigure}{r}{0.4\textwidth}
    \centering
    \vspace{-0.2in}
    \includegraphics[width=0.4\textwidth]{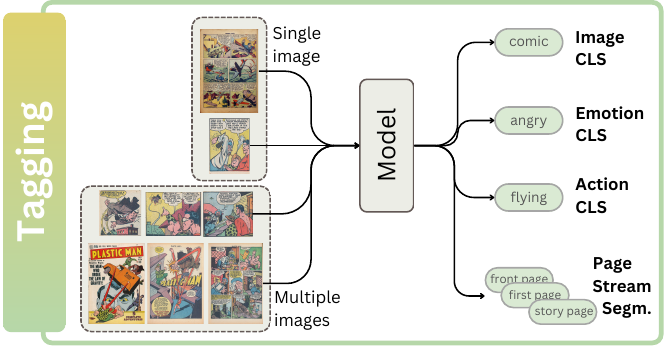}
    \caption{Illustration of \textit{Tagging} tasks.}
    \vspace{-0.2in}
    \label{ill:tagging}
\end{wrapfigure}

Tagging tasks predominantly involve classification outputs, albeit with varying input types (Fig. \ref{ill:tagging}).

\vspace{2mm}
\subsubsection{Image Classification}~\\%
\textbf{Definition:} The task of \textit{Image Classification} requires classifying an image into one of several predefined categories. In the domain of comics, the task can be applied at the panel or page level. In both cases, the classes vary from artist names, comic/manga styles, or image type (whether it's an ad, a front page, or a story page).

Early efforts in this domain, such as those by Chu et al. \cite{chu_linebased_2014_new}, focused on classifying manga panels by artistic style using Support Vector Machines (SVM) and manga-specific feature vectors. Their feature vectors are built on edge-detected lines creating a 20-dimensional vector from elements such as angle, orientation, the density of segments, etc. 
Hiroe et al. \cite{hiroe_histogram_2017} employed a novel method using SVMs to classify comic books based on the frequency of exclamation marks, demonstrating marks-per-page histogram as an innovative approach to content approximation.
Daiku et al. \cite{daiku_comic_2018} expanded the scope to page-level classification within manga stories using Convolutional Neural Networks (CNN), distinct from Page Stream Segmentation (discussed in Subsection \ref{subsec:pss}).
Jiang et al. \cite{jiang_learning_2018} introduced the Consensus Style Centralizing Auto-Encoder (CSCAE) for style classification, particularly distinguishing between Shonen (for boy) and Shojo (for girl) manga styles using robust style feature representations. They used low-rank and group sparsity constraints for consensus for the classification of manga character faces. Lastly, in \cite{terauchi_analysis_2019} Terauchi et al. presented an approach for classifying manga based on the author's unique style, employing a Variational Autoencoder (VAE) for this purpose. They also proposed a ``four-scene comics story dataset'' for this classification task into Mow, Seinen, and Shonen \textit{touches} (styles).
In a more recent work \cite{xu_panelpageaware_2023}, the authors tackled the challenge of (multi-)genre classification at the story level in Manga, introducing a Panel-Page-aware Comic genre classification model that takes page sequences as input and produces class-wise probabilities. Their proposal relyied on the attention mechanism to jointly attend page features and panel boxes, then pooled into a transformer encoder. The transformer classification token is merged with a Graph Convolution Network (GCN) over the labels graph, which predicts the probability distribution of the labels for the comic stories.

\vspace{2mm}
\subsubsection{Emotion Classification}~\\%
\textbf{Definition:} Emotion Classification in comics involves categorizing a single panel or an entire page (looking both at the image and the text) into predefined emotional categories, enhancing the understanding of the narrative's emotional context.

Tanaka et al. \cite{tanaka_relation_2016} conducted a seminal study on the relationship between speech balloon shapes and emotions in comics, using the Manga109 dataset to analyze how different balloon styles convey emotions.
The EmoReCon challenge at ICDAR 2021 \cite{nguyen_icdar_2021} pushed the field forward by tasking participants with classifying 8 emotions from comic panels using a specially designed dataset. One winning team employed a three-level fusion strategy (early, mid, and late), combining EfficientNetB3 \cite{tan_efficientnet_2020} for visual embeddings and RoBERTa \cite{liu_roberta_2019} for textual embeddings. Their mid-fusion approach outperformed other methods, showcasing the power of multimodal fusion. Another top team used an early fusion method, integrating ResNet50 image features with OCR-extracted text processed by a transformer encoder. They predicted emotions using a weighted average of cls tokens, achieving results comparable to the previous team.
Beyond character emotions, Sanches et al. \cite{sanches_manga_2016_new} analyzed readers' emotions by tracking physiological signals. By monitoring changes in the reader’s physiological responses, they classified manga into three genres: comedy, romance, and horror.
Yang et al. \cite{yang_manga_2023} introduced two new datasets, MangaAD+ and MangaEmo+, based on Manga109. MangaAD+ adds unannotated elements like shop names and ads, while MangaEmo+ includes annotations for 8 emotions on each page for multi-label classification. Using a modified Faster R-CNN for text detection and EfficientNet for visual features, they tackled the emotion classification task, though neither dataset has been made publicly available.

\vspace{2mm}
\subsubsection{Action Detection}~\\%
\textbf{Definition:} Action Detection in comics is characterized by the analysis of sequences of consecutive panels. This task involves processing multiple images to classify the depicted sequence into one of several predefined action categories.

Unique to Action Detection is its multimodal input nature, where each panel in the sequence contributes to a holistic understanding of the action. Unlike single image analysis, this task requires the model to interpret a series of images, each adding context and continuity to the narrative. Despite being popular for Videos \cite{zhu_comprehensive_2020}, this task has not yet been explored in comics apart from image-cloze task \cite{iyyer_amazing_2017} where the model is challenged to select the correct next panel from two options. In the context of clozure, many works have proposed valid solutions, starting from LSTM-based original solution of \cite{iyyer_amazing_2017}, the BERT-style multimodal and multigranularity embedding of \cite{Soykan2024ComicBERTAT}, and the Multimodal T5-based approach of \cite{vivoli_multimodal_2024}. However, while elementary, this task underscores the potential for more sophisticated Panel sequence analysis (and Action Detection) in comics, a domain where sequential imagery plays a pivotal role. Recently, \cite{Ikuta2024MangaUBAM} propose panel-level benchmarks about text and character recognition, as well as visual-cloze tasks in manga-style comics benchmarking recent open-source MLLMs models.

\vspace{2mm}
\subsubsection{Page Stream Segmentation}\label{subsec:pss}~\\%
\textbf{Definition:} Page Stream Segmentation (PSS) in comics involves dividing a comic book into different sections or stories. In its simplest form, this entails categorizing pages with tags like cover, regular page, ads, or last page. In a more advanced form, PSS extracts key information such as story titles, authors, illustrators, prices, characters, and synopses.

Though PSS is not a new concept in document analysis, it has seen various methods evolve over time. Traditional approaches like Support Vector Machines (SVM) for single-page classification \cite{gordo_document_2013} have given way to Convolutional Neural Networks (CNNs) \cite{wiedemann_page_2018} and more recent attention-based multimodal fusion techniques \cite{demirtas_semantic_2022}, which have significantly improved the understanding of page layouts and structures in documents.
However, PSS remains largely unexplored in the comics domain, where the task is complicated by the unique structure of comics. A single volume may contain multiple interwoven stories, ads, and narrative text, creating a more complex segmentation challenge. This complexity requires more sophisticated methods for segmenting and understanding the diverse content of comic books.
An emerging area of interest in PSS for comics is not just segmenting the pages but also tagging each segment with metadata such as story titles, authors, and synopses. 

\begin{wrapfigure}{r}{0.4\textwidth}
    \centering
    \vspace{-5mm}
    \includegraphics[width=0.4\textwidth]{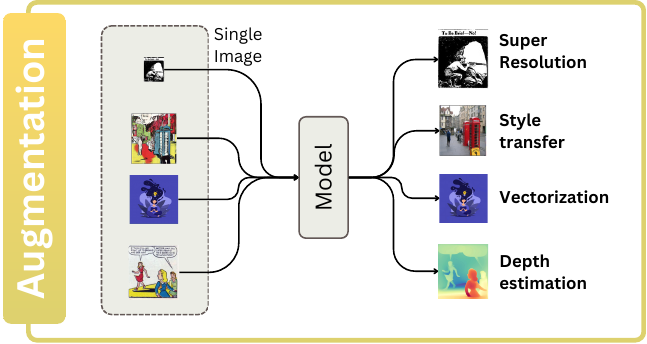}
    \vspace{-8mm}
    \caption{Illustration of \textit{Augmentation} tasks.}
    \vspace{-8mm}
    \label{ill:aumentation}
\end{wrapfigure}
This expands traditional image classification into a more detailed, story-oriented approach, offering a promising field of study.

\subsection{Augmentation}
\label{subsec:augmentation}

Augmentation tasks, in the context of comics, involve image manipulation techniques aimed at enhancing or altering the visual presentation  (Fig. \ref{ill:aumentation}). The primary tasks identified under this category are Super-Resolution (SR), Style-Transfer (ST), Vectorization, and Depth Estimation. Each of these tasks revolves around processing a single image input to produce an enhanced or stylistically altered output.

\vspace{2mm}
\subsubsection{Super-Resolution}~\\%
\textbf{Definition:} The Super-Resolution task in comics primarily targets the enhancement of image quality, especially for images that are compressed, photograph-based, or scanned. 

Recent advancements in manga super-resolution (SR) focus on preserving visual integrity during enhancement. Yao et al. \cite{yao_screentoneaware_2023} employ deep learning to classify and tailor SR models for manga screentone, ensuring semantic preservation. Dai et al. \cite{dai_structured_2022} introduce the Structured Fusion Attention Network (SFAN), which optimizes feature extraction and reconstruction quality through attention modules, achieving superior performance on datasets like Manga109. These methods enhance image quality while maintaining the artistic nuances critical to manga, effectively balancing technical efficiency with artistic integrity.

\vspace{2mm}
\subsubsection{Style-Transfer}~\\%
\textbf{Definition:} Style-transfer in comics encompasses a range of image-to-image modifications, including photo-to-comic transformation, image-to-image translation, and colorization, characterized by similar input and output attributes.

\textbf{Photo-to-Comic Transformation:}
Photo-to-comic has mainly involved mangas and involves replacing colors and textures with halftone patterns. Traditionally, it often consists of manually selecting and adjusting the colors. Methods have evolved from basic pattern generation to sophisticated techniques capturing structural and tonal similarities. Classic approaches relied on algorithmic solutions and were limited in their ability to handle multiple screentone. Recent advancements by Zhang et al. \cite{zhang_generating_2021_new} utilize deep learning for semantic region segmentation in cartoon images, followed by screentone application. However, these methods are primarily effective for color-based inputs, with limited applicability to line-based images.

\textbf{Image-to-Image Translation:}
Topal et al. \cite{topal_dassdetector_2023} explored style transfer as a transfer for the detection task. Specifically, they used style transfer networks like CycleGAN and CartoonGAN to train a detection network, which was then fine-tuned on specific comic datasets for character and face detection. This approach -using style transfer techniques to enhance performances in other tasks- has also been explored in different applications \cite{inoue_crossdomain_2018, deng_unbiased_2021}. In particular, the multi-task learning method for dense predictions in comic panels has been used as a medium for comic panel reconfiguration \cite{bhattacharjee_dense_2023} and depth estimation \cite{bhattacharjee_estimating_2022}.

\textbf{Colorization:}
Colorization in comics, particularly manga, involves transforming black-and-white images into full-color pages, requiring both artistic precision and technical expertise. Furusawa et al. \cite{furusawa_comicolorization_2017} introduced a method that allows users to add color dots to guide an automated process, giving users the flexibility to influence the final output while ensuring the system aligns with their vision. 
Hensman et al. \cite{hensman_cganbased_2018} used Conditional GANs (cGANs) to colorize manga panels by tailoring the model to each reference image, allowing versatile style adaptations but requiring a new GAN for every image, increasing computational demands. Tsubota et al. \cite{tsubota_synthesis_2019} explored 
\begin{wrapfigure}{r}{0.5\textwidth}
    \centering
    \vspace{-2mm}
    \begin{minipage}{0.23\textwidth}
        \includegraphics[width=\textwidth]{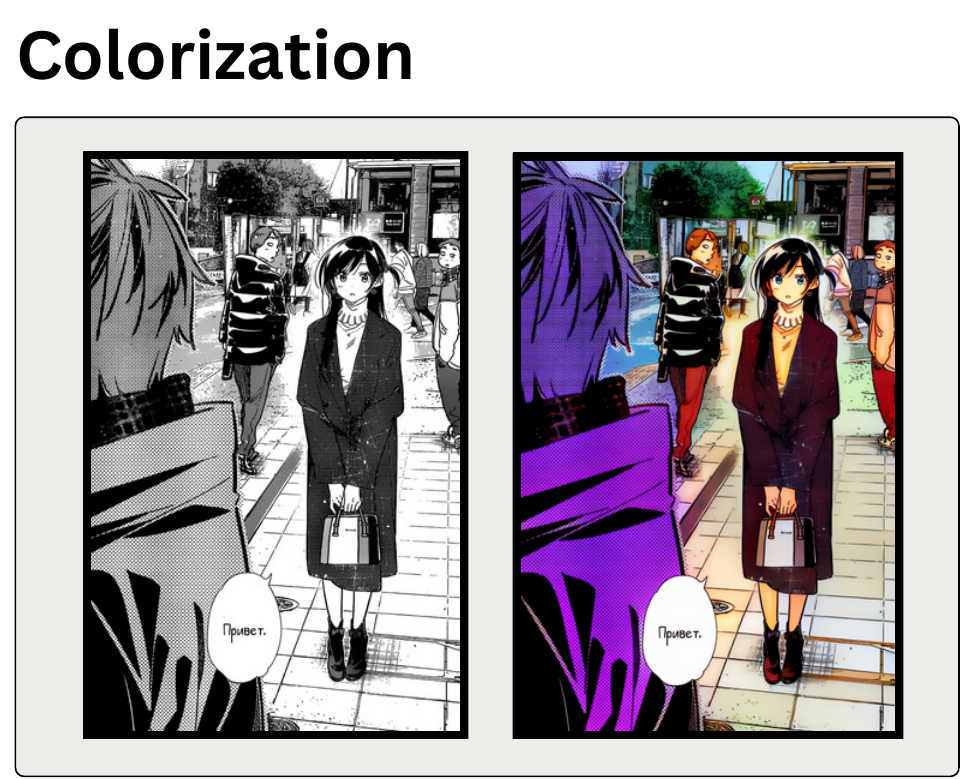}
    \end{minipage}\hfill
    \begin{minipage}{0.23\textwidth}
        \includegraphics[width=\textwidth]{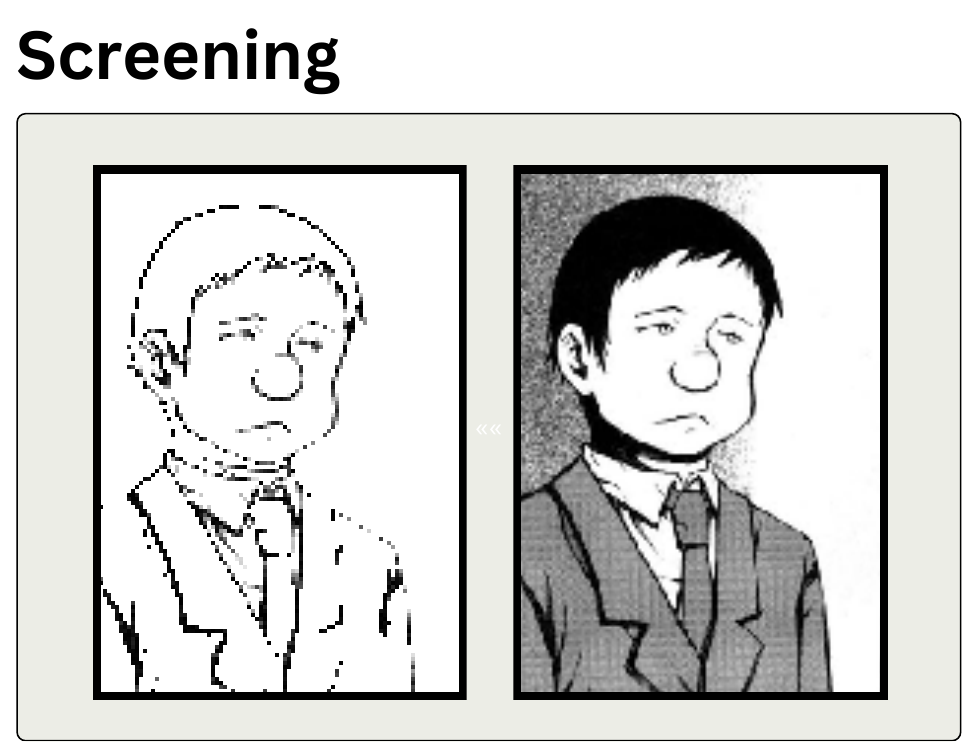}
    \end{minipage}
    \vspace{-2mm}
    \caption{Colorization vs. Screening}
    \label{fig:colo-screen}
    \vspace{-8mm}
\end{wrapfigure}
screentone synthesis, essential for manga, by comparing image translation techniques like pix2pix \cite{isola_imagetoimage_2018} on the Manga109 dataset. Their approach, using a U-Net architecture \cite{li_deep_2017}, stands out by explicitly considering screentone labels, preserving the screentone patterns integral to manga art.
Golyadkin et al. \cite{golyadkin_robust_2023} introduced a two-stage method for manga page colorization: Pixel2Style2Pixel \cite{richardson_encoding_2021} for color drafts, followed by a conditional GAN to refine the output. This method elegantly combines user-provided color hints and draft colorizations, yielding high-quality results that surpass previous methods. Jiramahapokee et al. \cite{jiramahapokee_inkn_2023} further refined the process using a multi-encoder Variational Autoencoder (VAE) to ensure color consistency and integrated CIELAB interpolation \cite{zeyen_color_2018} to enhance color saturation and authenticity. Their approach produced detailed and visually consistent colorizations, particularly in shading quality.

\textbf{Screening:}
Screening, or adding screentone to black-and-white manga drawings, has been a focus of several studies. A notable contribution by Wu et al. \cite{wu_shadingguided_2023} introduced a two-stage architecture for this process. The first stage involves generating grayscale shading using a six-layer Swin-transformer \cite{liu_swin_2021}, which then serves as input for the second stage, a reference-based screentone generation module. This algorithmic process extracts screentone patterns from a reference manga and applies them to line drawings, considering the shading nuances provided by the first stage.

These advancements in colorization represent a significant contribution to the domain of comics, particularly manga, where colorization can profoundly impact the narrative and aesthetic appeal of the stories.

\vspace{2mm}
\subsubsection{Vectorization}~\\%
\textbf{Definition:} Vectorization in the context of comics, particularly manga, involves transforming raster images into vector formats. This process includes operations like screening, structural lines extraction, and converting images into vector graphics, each with unique methodologies and applications.

\textbf{Structural Lines Extraction:}
This process, structural lines extraction, focuses on removing screentones to leave only the defining lines of figures and shapes. It can be seen as the reverse operation of the screening, which corresponds to adding the screentone to a structural line image. In this task, Li et al. \cite{li_deep_2017} developed a CNN-based method using a U-Net architecture with residual blocks to effectively extract structural lines from manga pages. This method is adaptable for various applications, including manga retargeting and colorization, and can handle diverse patterns and textures, resulting in clean skeleton sketches.

\textbf{Manga Vectorization:}
Various works \cite{kopf_depixelizing_2011, selinger_potrace_2003, ma_layerwise_2022} have proposed different approaches for this task. In particular, Yao et al. \cite{yao_manga_2017} developed a method involving adaptive binarization and screentone detection, refining borders, estimating lighting, and compensating for missing strokes, leading to high-quality, resolution-independent rendering. A recent innovative approach, MARVEL by Su et al. \cite{su_marvel_2023}, employs Reinforcement Learning to train an agent in selecting drawing primitives that recreate raster manga images in vector format. While effective, this approach works with squared manga patches and may result in visual gaps in certain renderings. Moreover, this method is not convenient in shape rendering, as it uses primitives and cannot manipulate them. MARVEL represents an interesting exploration into the use of machine learning for vectorization, highlighting both the potential and the complexities of this task.

\vspace{2mm}
\subsubsection{Depth Estimation}~\\%
\textbf{Definition:} Depth Estimation within the comic domain refers to the process of predicting the spatial distance between objects and the viewer in a two-dimensional comic panel. It serves a crucial purpose in the \textit{reconfiguration} of comic panels and unlocks potential developments in digital artistry. 

To date, this need has arisen primarily when adapting comics for digital media like smartphones, tablets, and e-readers, where the original layout may not fit the screen optimally. Key methods for reconfiguring panels include: (i) Adjusting panel proportions and rescaling them to fit the new medium while maintaining narrative and visual integrity, and (ii) Cropping panels to preserve essential elements. Both approaches rely on depth estimation to identify and prioritize key components.
A major challenge in this area is the lack of ground truth data for comics. To address this, Bhattacharjee et al. \cite{bhattacharjee_estimating_2022} employed Image-to-Image style-transfer, transforming comic images to resemble natural ones, enabling depth estimation using Laplacian edges and feature-based GANs. Building on this, a follow-up study \cite{bhattacharjee_dense_2023}, based on their earlier work \cite{bhattacharjee_mult_2022}, introduced a Swin Transformer for multitasking, including depth estimation and object segmentation. However, the multi-decoder-head architecture poses scalability challenges due to task complexity.
Depth estimation is crucial for adapting comics to various digital formats, ensuring key visual elements are retained while preserving the original artistic intent.

\subsection{Satellite Tasks}
Some of the tasks that fit under the Augmentation umbrella have not found a space in our taxonomy, e.g. the task of de-warping. The task involves correcting distortions typically found in scanned comic pages. Addressing the de-warping challenge, a recent study by Garai et al. \cite{garai_automatic_2023} proposed a novel mathematical model to describe the warping process. This model estimates distortion factors based on the boundaries of panels in a comic document image, aiming to rectify these warps effectively. While their approach yielded promising qualitative results, particularly in the image areas, challenges persist in handling text distortion, where some distortions proved unrecoverable.

\section{Layer 2: Grounding, Analysis, and Segmentation}
\label{sec:locu-layer2}

The second layer of our \textit{LoCU} framework comprises three distinct task groups: \textbf{grounding}, \textbf{analysis}, and \textbf{segmentation}. These tasks dig deeper into the intricate components of comics, focusing on detailed elements like panels, text, and characters and their ordering and associations.

\subsection{Grounding}
\begin{wrapfigure}{r}{0.4\textwidth}
    \centering
    \vspace{-0.2in}
    \includegraphics[width=0.4\textwidth]{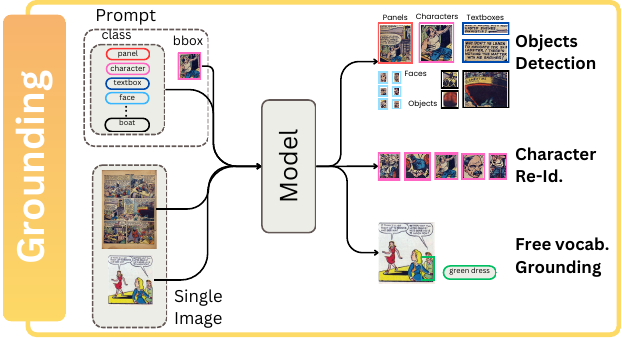}
    \caption{Illustration of \textit{Grounding} tasks.}
    \vspace{-0.2in}
    \label{ill:grounding}
\end{wrapfigure}
Grounding in comics involves identifying and classifying specific elements within a comic's panel, such as text, characters, and objects (Fig. \ref{ill:grounding}). Under this task name umbrella, many individual tasks belong such as Detection, Character Re-identification, and sentence Grouding.

\vspace{2mm}
\subsubsection{Detection}~\\%
\textbf{Definition:} In \textit{Detection} task, given an image and a set of class names (or tags), the goal is to find all instances of those classes in the image and locate them with bounding boxes.

\textbf{Panels.}
Initial studies in digital Manga, such as those by Ponsard et al. \cite{ponsard_enhancing_2009} and Arai et al. \cite{arai_method_2010}, emphasized automated segmentation and panel detection, underscoring the importance of maintaining reading order. Arai et al. \cite{arai_method_2011} further developed these ideas with a focus on real-time detection and extraction techniques, which evolved into more sophisticated SVG-based approaches as explored in \cite{su_recognizing_2011}. \cite{ho_panel_2012} contributed advanced methods for panel extraction using region growing techniques and mathematical morphology, combined with speech balloon detection.

\textbf{Text.}
Jomaa et al. \cite{jomaa_panel_2022_new} employed a tracking algorithm for panel extraction, while speech balloons were identified using Robert's edge detection operator and classified with thresholding. Liu et al. \cite{liu_textaware_2016} focused on balloon extraction based on text character locations. Pal et al. \cite{pal_linewise_2016} performed text extraction using SIFT local features, Space Pyramid Matching (SPM), and SVM on eBDtheque and Bangla comics images. In \cite{aramaki_text_2016} authors contributed to text extraction in Manga109 combining region proposal generation with SVM classifiers. Chu et al. \cite{chu_text_2018} and Qin et al. \cite{qin_faster_2017} modified existing architectures like Faster R-CNN for detecting text and faces in comics. In particular, \cite{chu_text_2018} proposed additional aspect ratio regions, while \cite{qin_faster_2017} built two comics datasets, namely JC2463 (Manga black and white style) and AEC912 (colored American comics style) to test face detection evaluation.

\textbf{Face.}
Interestingly, in \cite{yanagisawa_study_2018} authors benchmarked Fast R-CNN, Faster R-CNN and SSD in panel, character, face, and balloon detection. They used the Manga109 dataset, discovering that Fast R-CNN works better in panels when there are clear boundaries, Faster R-CNN when boundaries are not well defined, and SSD needs already cut panels as it is limited for full-page complex mangas.
So far, many methods have used classical detection architectures and applied them to the comic domain to extract elements such as panels, characters, text, etc. A new direction is drawn by \cite{nguyennhu_what_2019} challenged the conventional four-edge polygon representation of comic panels, suggesting segmentation-based approaches using U-Net for more accurate extraction. Here, authors performed extraction through segmentation, employing a U-Net that classifies pixels into background, panels, and borders.

\textbf{Onomatopoeia.}
Recently, the focus has shifted to onomatopoeia detection, a relatively unexplored area in comic analysis. The release of the COO dataset by Baek et al. \cite{baek_coo_2022} has paved the way for new research in this domain. Authors proposed a modified version of M4C \cite{hu_iterative_2020} designed to link truncated onomatopoeias together. Following this, Yang et al. \cite{yang_manga_2023} proposed text augmentation techniques for onomatopoeias using EfficientNet.
Sharma et al. \cite{sharma_cpd_2023} furthered this research by introducing their dataset for panel and character detection, employing a Faster R-CNN model. However, they did not provide comparative analysis with other models.

\textbf{Benchmarks.}
Lately, Dutta et al. \cite{dutta_cnn_2019, dutta_bcbid_2022} introduced the BCBId dataset, benchmarking existing datasets (eBDtheque, Manga109, and DCM) and employing YOLO architectures for panel and character detection. They presented comparisons with existing methods, showing enhanced performance in detecting panels and characters across multiple datasets. However, it is unclear if these results are truly comparable. This is due to the fact that in previous (and contemporary) works, authors don't provide neither information about the settings used (in terms of train-val-split), nor about hyperparameters and model settings to replicate the weights. 
Only recently, Vivoli et al. \cite{vivoli_cdf_2024} fairly compare models across various datasets and styles, unifying annotations for multiple detection classes and providing model weights, code, and data to replicate.

\vspace{2mm}
\subsubsection{Character Re-Identification}~\\%
\textbf{Definition:} \textit{Character Re-Identification}, also known as character retrieval in comics analysis, has evolved from focusing on visual features alone to incorporating textual information alongside character images.

Initial studies by Khan et al. \cite{khan_color_2012} and Amirshahi et al. \cite{amirshahi_phog_2012_new} employed techniques like PHOG and SIFT keypoints for character retrieval. Iwata et al. \cite{iwata_study_2014}, inspired by Sun et al. \cite{sun_similar_2011}, adapted these methods for manga characters, focusing on character-level identification rather than entire page regions, and further explored their applications in \cite{iwata_similarity_2015}. However, Li et al. \cite{li_dual_2020} highlighted the limitations of these handcrafted features, particularly their sensitivity to pose, expression, and shape variations.
Li et al. \cite{li_dual_2020} addressed manga character retrieval and verification, the latter being the task of classifying whether two images belong to the same character. They introduced a dual loss approach—combining dual ring loss and dual adaptive re-weighting loss—to address long-tailed distributions and quality variations in manga data, improving robustness.
In animated cartoons, where characters frequently change shape and color, Nir et al. \cite{nir_cast_2022} proposed a Self-Supervised Learning (SSL) and Multi-object Tracking (MOT) approach to cluster and link character proposals, facilitating character re-identification across scenes. Similarly, Zhang et al. \cite{zhang_unsupervised_2022} tackled the challenges of character re-identification in manga, where faces are similar but body features differ. They introduced FSAC (Face-body and Spatial-temporal Associated Clustering), which refines clustering through face-body graphs and spatial-temporal relationships, using a triplet loss to handle artistic exaggerations.
To address occlusion in manga, Zhang et al. \cite{zhang_occlusion_2024_new} developed Occlusion-Aware Manga Character Re-identification (OAM-ReID), using self-paced contrastive learning. Their model synthesizes occluded data, such as speech balloons and incomplete bodies, and combines contrastive, occlusion, and re-identification losses, training a ViT and TransReID on 10 Manga109 books.
Soykan et al. \cite{soykan_identityaware_2023} introduced "Identity-aware SimCLR" for character re-identification, leveraging SimCLR \cite{chen_simple_2020} with data augmentations to capture both local (face) and global (body) features. Similarly, Sachdeva et al. \cite{sachdeva_manga_2024} proposed Relationformer \cite{shit_relationformer_2022}, a multi-task Transformer model for character clustering and linking bounding boxes at the page level.
Finally, Ahmad et al. \cite{ahmad_crossdomain_2023} introduced an enhanced YOLOv5 combined with XGBoost for object detection, achieving state-of-the-art performance across multiple datasets, emphasizing the importance of well-annotated data and augmentation techniques in improving detection tasks.



\vspace{2mm}
\subsubsection{Sentence-based Grounding}~\\%
\textbf{Definition:} Sentence-based grounding is a Vision-Language task that links textual content to specific visual elements within a panel. In comics, this task translates into grounding the sentence elements both into a single panel and a full page. Moreover, the complexity of the task varies depending on what is the sentence to be grounded: a panel description or a dialogue or narrative description.

In image and video Multimedia processing various works tackle the task with custom CNNs and LSTMs \cite{li_zeroshot_2019} approaches, in an unsupervised manner \cite{parcalabescu_exploring_2020} or with custom multimodal fusion approaches \cite{shen_groundvl_2023,liu_grounding_2023}.
However, the only work that used a similar approach (GroundingDINO) in a zero-shot setting is \cite{sachdeva_manga_2024} which used it to automatically annotate the training set.

\subsection{Analysis}
\label{subsec:analysis}
\begin{wrapfigure}{r}{0.4\textwidth}
    \centering
    \vspace{-0.2in}
    \includegraphics[width=0.4\textwidth]{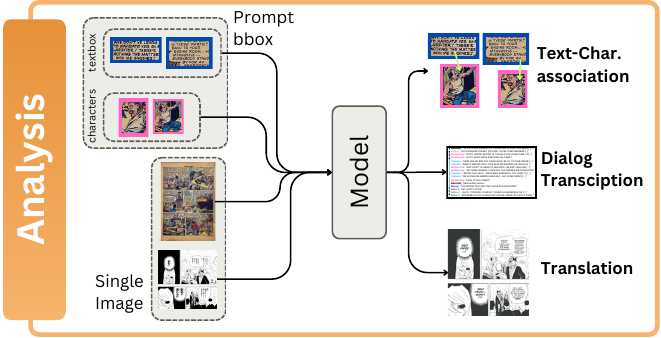}
    \caption{Illustration of \textit{Analysis} tasks.}
    \vspace{-0.2in}
    \label{ill:analysis}
\end{wrapfigure}
Analysis tasks involve a deeper examination of the relationship between text and visual elements in comics, including character-text association, panel sorting, dialog transcription, and translation (Fig. \ref{ill:analysis}).

\vspace{2mm}
\subsubsection{Text-character association}~\\%
\textbf{Definition:} The task of \textit{text-character association} in comics aims to link textual content, usually dialogue, to the corresponding speaking character. This task often involves a two-step process: first identifying panels, text boxes, and character boxes, and then associating the text with its speaker.

Rigaud et al. \cite{rigaud_speech_2015} addressed the text-character association task using geometric graph analysis and anchor point selection, developing a distance-based algorithm that utilized both global (page-level) and single-frame (panel-level) information, relying on Euclidean distances and balloon tail cues.
Building on this, Nguyen et al. \cite{nguyen_comic_2019} introduced ComicMTL, a multitask learning approach based on a modified Mask R-CNN with an added "PairPool" head for association prediction. Their method tackled multiple tasks—panel and character detection, balloon segmentation, text recognition, and text-speaker association—demonstrating effectiveness on DCM772 and eBDtheque datasets.
Omori et al. \cite{omori_algorithms_2022} took a novel approach by focusing on the continuity of comics, particularly reading order, to associate text with off-panel speakers. However, their approach raised scalability and generalizability concerns. Li et al. \cite{li_manga109dialog_2023} expanded the Manga109 dataset by introducing the Manga109 Dialog dataset, refining speaker detection. They categorized text-character links into “easy” (character in the panel) and “hard” (character off-panel), employing Faster R-CNN for detection and association classification. Despite progress, handling alternating speaker dialogue remains challenging, pointing to the potential of NLP models for complex scenarios.
Sachdeva et al. \cite{sachdeva_manga_2024} recently proposed a multi-task Transformer model, Relationformer \cite{shit_relationformer_2022}, with specialized MLP heads for detection and linking. Their two-step, end-to-end solution first detects and then links characters and text, though their model's benchmarking, especially against YOLO-based models, remains limited, focusing only on GroundingDino \cite{liu_grounding_2023} in a zero-shot setting.
Lastly, in \cite{Soykan2024SpatiallyAS} built on top of Comic-MTL \cite{nguyen_mangammtl_2021} a Mask R-CNN based model for detecting and linking textboxes and characters, but did not fully compared with previous models.

Character-text association is a crucial yet challenging task in comics analysis, requiring an understanding of narrative and visual-textual interplay. Future research could benefit from advanced NLP techniques to manage complex dialogue scenarios, and more unified benchmarks and replicable methodologies are needed, as highlighted by the \textit{LoCU} framework, to support comparative analysis and innovation in this field.

\vspace{2mm}
\subsubsection{Panel Sorting}~\\%
\textbf{Definition:} \textit{Sorting} individual comic panels into their correct narrative order presents a unique challenge, both for humans and automated systems.

Ueno et al. \cite{ueno_story_2018} conducted a study revealing that while people can relatively easily identify the first and last panels in a sequence, accuracy diminishes with the presentation of all panels from four-panel strips. This suggests the complexity involved in discerning narrative continuity in comics.
The same study attempted to automate panel sorting using an AlexNet-like architecture. However, the CNN’s performance was poor, indicating the difficulty of the task and the need for more advanced or specialized algorithms in panel sorting.

\vspace{2mm}
\subsubsection{Dialog transcription}\label{sec:dialog}~\\%
\textbf{Definition:} \textit{Dialog transcription} in comics aims to create an automatic transcript of dialogues, identifying the speaker and the sequence of speech.

The process typically involves several steps: (i) detecting panels, text boxes, and character boxes, (ii) associating text with speakers, and (iii) sorting text boxes according to their reading order to generate a coherent dialogue transcript.
This task has been addressed by \cite{li_manga109dialog_2023} for Japanese Mangas, proposing the needed annotations for Manga109, and more recently by \cite{sachdeva_manga_2024} for English Mangas called PopManga. However, a few limitations arise from these approaches: sorting the text boxes is always done as a postprocessing operation, depending only on algorithmic approaches. Cases in which balloons alternate among close panels are inherently mistaken. Recently, Sachdeva et al. \cite{sachdeva_tails_2024} proposed Magiv2 which tackled the problem of dialog transcription adding also names (from a name bank) allowing it to operate on the whole collection of pages.
In CoMix \cite{vivoli_comix_2024}, a first benchmark for this task was proposed, authors enhanced the previous annotations and proposed a new metric for character naming (based on ANLS) and dialog transcription (based on Hungarian matching and edit distance).

\vspace{2mm}
\subsubsection{Translation}~\\%
\textbf{Definition:} The task of translation in comics involves not only linguistic challenges but also graphical and spatial ones, requiring an integrated approach that respects the visual and narrative structure of the comic. Text appears in various forms such as balloons, onomatopoeias, and scene text, each necessitating different strategies. For instance, \textit{Balloon Text} must fit within the original speech balloon, preserving visual integrity, while \textit{Onomatopoeias} may require both linguistic translation and graphic replacement to maintain stylistic consistency.

Recently, several works have addressed these complexities by integrating multimodal information and context-aware techniques. \cite{Hinami2020TowardsFA} were among the first to incorporate context obtained from manga images into a multimodal translation framework, developing a comprehensive system for fully automated manga translation. To improve text image translation performance, \cite{Ma2023MultiTeacherKD} proposed a Multi-Teacher Knowledge Distillation method, effectively distilling knowledge from a pipeline model into an end-to-end Text Image Machine Translation (TIMT) model. The advent of Large Language Models (LLMs) has further advanced the field; \cite{YangLargeLM} explored leveraging LLMs for manga translation tasks, demonstrating the potential of these models in handling complex scenarios involving both text and visual elements. Similarly, \cite{Lippmann2024ContextInformedMT} proposed a methodology that leverages the vision component of multimodal LLMs to improve translation quality, investigating the impact of translation unit size and context length, and introducing a token-efficient approach along with a new evaluation dataset. Inspired by human translation workflows, \cite{Singh2024TheFO} presented an approach to completely automate the process of translating graphic novels, addressing the challenges of automating different aspects of the translation workflow. Focusing on character identification and speaker prediction, \cite{Li2024ZeroShotCI} proposed an iterative multimodal framework that employs multimodal information for both tasks, allowing machines to identify characters and predict speaker names based solely on unannotated comic images. Furthermore, \cite{Liu2024GeneratingVS} introduced the task of character-centric story generation, presenting the first model capable of generating visual stories with consistently grounded and coreferent character mentions, emphasizing the importance of character consistency and coreference in narrative understanding and translation. Despite these tasks being tackled as image-to-text translation, they have origins in image-to-image translation due to possible changes in visual components influenced by cultural elements, as discussed by \cite{khanuja2024image}. This indicates that future research may focus more on end-to-end image-to-image translation and transcreation, where both textual and visual elements are adapted to enhance cultural relevance.

\subsection{Segmentation}
\begin{wrapfigure}{r}{0.4\textwidth}
    \centering
    \vspace{-0.2in}
    \includegraphics[width=0.4\textwidth]{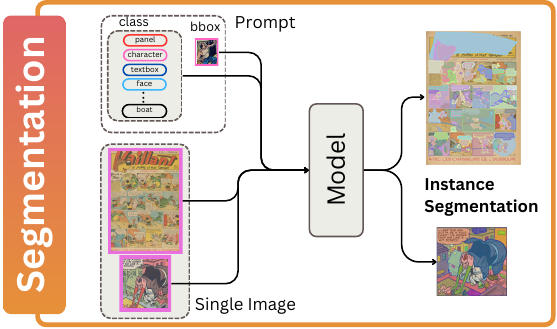}
    \caption{Illustration of \textit{Segmentation} tasks.}
    \vspace{-0.2in}
    \label{ill:segmentation}
\end{wrapfigure}
Segmentation in comic analysis, particularly instance segmentation, presents a nuanced approach to identifying and isolating various elements within a comic panel (Fig. \ref{ill:segmentation}).

\vspace{2mm}
\subsubsection{Instance segmentation}~\\%
\textbf{Definition:} \textit{Instance segmentation} in comics extends beyond the scope of typical object detection. Rather than just identifying and classifying elements within bounding boxes, this task involves generating precise pixel-wise masks for each instance.

Instance segmentation, as opposed to detection, offers more accurate panel separation and representation in comics, which is crucial when reconfiguring or reformatting them for different layouts. This precision ensures that each panel is cleanly isolated, avoiding remnants of adjacent panels or the original layout—a common issue with less precise bounding boxes.
One of the key challenges in instance segmentation for comics lies in the diversity of artistic styles and the complexity of layouts. Comics often feature intricate designs, overlapping elements, and various configurations such as grids, staggered layouts, bleeds, and insets, as noted by Cohn \cite{cohn_visual_2023}. These complex panel arrangements make it difficult to segment panels accurately using bounding boxes, especially when panel boundaries are not clearly defined.
In a recent work \cite{Kouletou2024InvestigatingNN}, authors proposed a pipeline of detection models (YOLO) and segmentations (GroundingDINO) for segmenting objects in comic panels. Despite the idea being simple and elegant -using in a cascade approach a YOLO detector and prompting SAM with the detected bounding boxes- the approach has been only evaluated for the detection tasks as no Segmentation ground truth exists for the used datasets. Moreover, as noticed in \cite{vivoli_cdf_2024}, results are not comparable due to different evaluation datasets/splits/criteria.

Although current datasets do not include semantic segmentations, the 2024 ``AI for Visual Art'' workshop \cite{bhattacharjee_github_2024} introduced a challenge on comic semantic segmentation, signaling the growing importance of this task in comic analysis.

\section{Layer 3: Retrieval and Modification}
\label{sec:locu-layer3}

Layer 3 in the \textit{Layers of \comicsunderstanding} focuses on advanced image and text understanding through tasks such as Retrieval (uni-modal and multi-modal) and Modification. 

\subsection{Retrieval}
\label{sec:retrieval}
\begin{wrapfigure}{r}{0.4\textwidth}
    \centering
    \vspace{-0.2in}
    \includegraphics[width=0.4\textwidth]{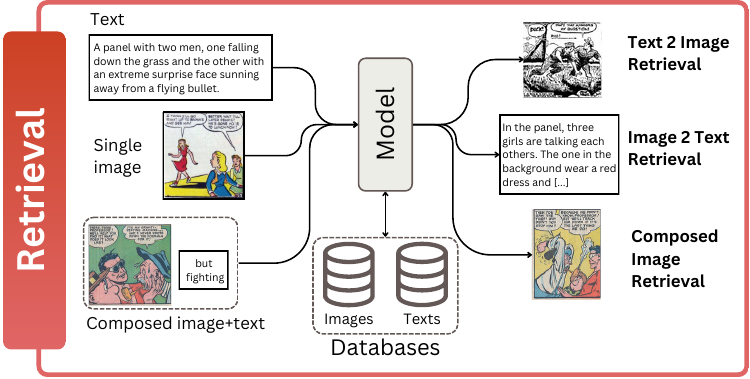}
    \vspace{-5mm}
    \caption{Illustration of \textit{Retrieval} tasks.}
    \vspace{-5mm}
    \label{ill:retrieval}
\end{wrapfigure}
The task of retrieval, in the domain of the comic, addresses two main retrieval settings in: unimodal and cross-modal retrieval, which do not differ, in practice, from the vision-language retrieval tasks (Fig. \ref{ill:retrieval}). These tasks aim to retrieve an instance of the same modality of the query (unimodal) or a different modality from the query (cross-modal), using either text or images as the query. An additional task emerged recently in the realm of Vision-Language, called ``Composed image retrieval'', which combines in the query both image and text.

\vspace{2mm}
\subsubsection{Unimodal Retrieval}~\\%
\textbf{Definition:} Unimodal retrieval in comics involves searching for a specific media element (image or text) within a database, using a query of the same format (image or text, respectively).

Pioneering works in this area have explored various aspects of comic retrieval. Wu et al. \cite{wu_searching_2010} developed a comic retrieval system that leveraged OCR and external text.
One significant article \cite{sun_similar_2011} focused on retrieving full pages or books containing similar elements. This approach is particularly relevant in copyright violation contexts. The authors implemented feature vectors for each image, storing them in a database, and used these vectors to find the most similar documents. Their feature representation method was based on the Histogram of Oriented Gradients (HOG), which outperformed the traditional SIFT method. Notably, they applied HOG not to the entire images but selectively to Regions Of Interest (ROIs) identified in the comics, such as faces or specific markings.
Building on this approach, authors in \cite{iwata_study_2014} utilized a similar methodology for manga character retrieval. They focused on identifying and extracting distinctive features from characters, enhancing the ability to retrieve specific characters from large databases.

\vspace{2mm}
\subsubsection{Cross-modal retrieval}~\\%
\textbf{Definition:} Cross-modal retrieval in comics encompasses tasks where the query and the retrieval results are from different modalities, such as using sketches or text queries for retrieving a comic image.

\textbf{Sketch-based Retrieval.}
Matsui et al. \cite{matsui_sketch2manga_2014} introduced a framework for retrieving manga images from sketches using the Fine Multi-scale Edge Orientation Histogram (FMEOH). This method indexed different-sized squares on a page and enabled efficient retrieval from sketch queries.
Further advancing this area, authors \cite{matsui_sketchbased_2017} proposed a two-step system that first processes manga and then retrieves images based on sketches. This framework was one of the first to employ the Manga109 dataset and included margin labeling, objectness-based edge orientation histograms, and approximate nearest neighbor search. Narita et al. \cite{narita_sketchbased_2017} explored CNN-based feature extraction for sketches and manga images (without and with screentone, respectively), demonstrating improvements over algorithmic approaches.

\textbf{Multimodal Learning and Character Retrieval.} 
Nguyen et al. \cite{nguyen_mangammtl_2021} proposed a multitask multimodal approach for character image and text learning, aiming at tasks such as character retrieval and emotion recognition. Their method focused on leveraging both visual and verbal information in manga content, in a CLIP-like style aiming at performing various character tasks (e.g. detection, retrieval, emotion classification), trained uniquely with visual information and verbal information in manga image content.
Later, in \cite{wei_comiclib_2022} authors introduced ComicLib, a dataset for comic sketch research, providing benchmarks across various tasks like colorization, generation, and retrieval. They conducted extensive comparison experiments with other datasets (e.g. QuickDraw \cite{ha_neural_2017}) to provide a benchmark for ComicLib on common tasks like colorization (Style2Paints \cite{zhang_style_2017}), generation (DCGAN \cite{radford_unsupervised_2016}), retrieval (image-to-image with ResNet), detection (with Faster R-CNN \cite{ren_faster_2016}, YOLOv3 \cite{redmon_yolo_2016}, and SSD \cite{liu_ssd_2016}) and recognition. However, the dataset’s availability is unclear.

\textbf{Text-to-Image Retrieval.} 
Shen et al. \cite{shen_maru_2023_new} presented a system for text-to-image content retrieval within Manga frames. This multi-staged system integrates object detection, text recognition, and a vision-text encoder to facilitate efficient search and retrieval of dialogues and scenes. Their method is composed of an object detection model for identifying text and frame bounding boxes (DETR-based \cite{carion_endtoend_2020}), a Vision Encoder-Decoder model for text recognition (DiT \cite{touvron_training_2020}), a text encoder for embedding text (multilingual DistilBERT \cite{sanh_distilbert_2020}), and a vision-text encoder with unified embedding space (CLIP \cite{radford_learning_2021}). They experimented with Japanese and GPT-4\footnote{Using the website \href{https://platform.openai.com/}{https://platform.openai.com/}}translate/rephrased English sentences, on the annotated panels from ``Dollgun Book''. Despite being a composition of existing systems on panel level, on a small distribution of data, the work is a first direction to more complex Comic retrieval.

\subsection{Modification}
\begin{wrapfigure}{r}{0.4\textwidth}
    \centering
    \vspace{-0.2in}
    \includegraphics[width=0.4\textwidth]{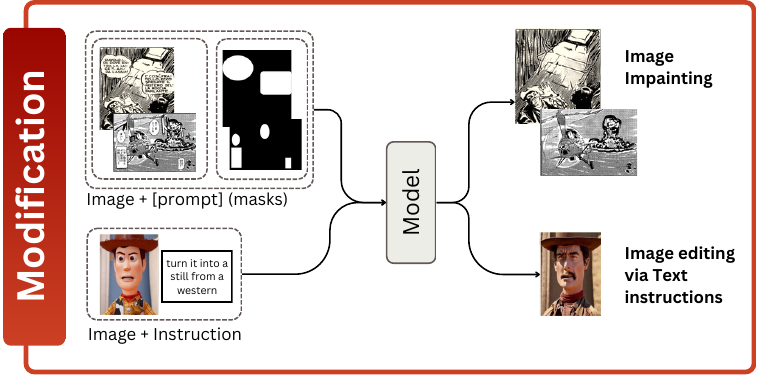}
    \vspace{-5mm}
    \caption{Illustration of \textit{Modification} tasks.}
    \vspace{-8mm}
    \label{ill:modification}
\end{wrapfigure}
In the realm of comics, modification tasks such as image inpainting and image editing play a crucial role in streamlining the storytelling process and enhancing creative expression (Fig. \ref{ill:modification}).

\vspace{2mm}
\subsubsection{Image Inpainting and Editing}~\\%
\textbf{Definition:} Image inpainting and editing in the context of comics involve various techniques and tools that allow for the alteration or enhancement of comic panels, facilitating more dynamic and engaging storytelling.

In the realm of comic inpainting and editing, CodeToon \cite{suh_codetoon_2022} offers a code-driven storytelling tool that automatically generates comics from code snippets through metaphorical interpretation. It simplifies the transition from story ideation to comic format, addressing common challenges faced by comic creators.
ComicScript \cite{wang_interactive_2022} provides a suite of operations for interactive comic creation, allowing users to add/remove panels, change perspectives, and support branching narratives. It also integrates data manipulation to enhance engagement. Professional creators validated its potential by crafting interactive comics, showcasing the tool's ability to personalize storytelling.
Both tools highlight the growing role of computational methods in enhancing comic creation and editing workflows.


\subsection{Future Tasks}

\vspace{2mm}
\subsubsection{Personalized Image retrieval}~\\%
\textbf{Definition:} Personalized Image Retrieval (PIR) focuses on developing person-centric models that can efficiently retrieve specific comic panels based on a compound query involving both image and text inputs \cite{korbar_personalised_2022}. PIR aims to identify and retrieve images or videos that correspond to a compound query, which typically includes an image of a person’s face combined with a textual scene or action description.

In the realm of comics, this translates to the capability of retrieving panels where a specific character appears, given an image of the character, the character's name, or a description of the character. This task becomes increasingly complex when the retrieval criteria include context-specific actions (e.g., a character eating pizza or jumping out of a window) or interactions with other characters.

\section{Layer 4: Advanced Visual-Language Understanding}
\label{sec:locu-layer4}
In the fourth layer, we propose in the \textit{Layer of Comics Understanding} a set of tasks related to multimodal understanding, which comprehends often reasoning through a text-rich image given a text prompt and one or more images as input. However, the only task that has been explored in comics already is Visual Question Answering, which under its umbrella groups different definitions of VQA in comics.

\subsection{Understanding}
\begin{wrapfigure}{r}{0.4\textwidth}
    \centering
    \vspace{-0.2in}
    \includegraphics[width=0.4\textwidth]{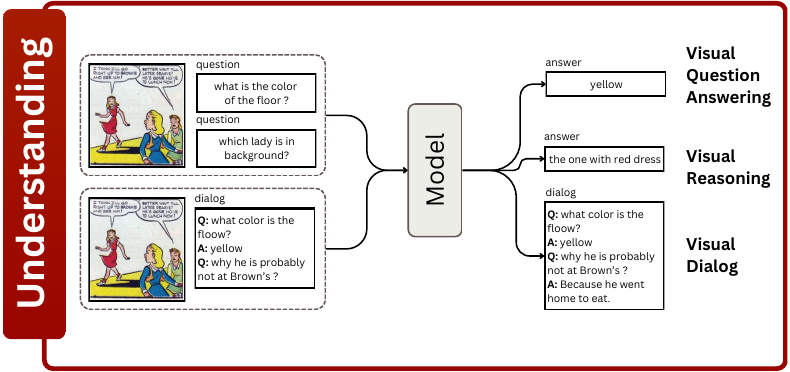}
    \vspace{-5mm}
    \caption{Illustration of \textit{Understanding} tasks.}
    \vspace{-5mm}
    \label{ill:understanding}
\end{wrapfigure}
This layer investigates tasks that require a deeper integration of visual elements and language, advancing beyond basic identification or retrieval to more complex forms of comprehension and reasoning within comics (Fig. \ref{ill:understanding}).

\vspace{2mm}
\subsubsection{Visual Question Answering}~\\%
\textbf{Definition:} Given an image-question pair, \textit{Visual Question Answer} requires answering a question based on the image. Most studies treat VQA as a classiﬁcation problem on a predeﬁned answer set, but its general definition would be an open answer where no specific set of options is provided.

The first notable attempt in the domain of the comic is by \cite{moriyama_designing_2016} with ComicVQA. However, their approach, more akin to retrieval based on specific ``Doraemon'' comics datasets, diverges from the typical VQA framework.
Later, Sumi et al. \cite{sumi_comicqa_2017} presented a system, ComicsQA, with a unique twist, converting user QA interactions into comic stories, reflecting the user's situation and offering solutions.
The creation of a comprehensive VQA dataset specifically for comics, encompassing single-panel, single-page, and multi-page formats, remains an open challenge. In a recent survey \cite{zeng_survey_2020}, Zeng et al. show the massive work that has been carried out on VQA on multi-modal data across different mediums (movies, comics, etc.) and Multimodal Machine Comprehension \cite{sahu_solving_2021, sahu_challenges_2022}. This, together with multiple datasets across various mediums \cite{singh_vqa_2019,biten_scene_2019,mathew_infographicvqa_2021,vivoli_mustvqa_2022,tanaka_slidevqa_2023,tanaka_slidevqa_2023} are laying the groundwork for advancing VQA also in comics.

\vspace{2mm}
\subsubsection{Visual Entailment}~\\%
\textbf{Definition:} Visual entailment (VE) in comics is an area yet to be deeply explored. This task involves assessing whether a given image semantically entails the accompanying text. The challenge lies in interpreting the visual narrative in conjunction with textual elements to ascertain semantic consistency or correlation. It can be seen as a simplified version of the text-closure task where we aim at correctly classifying whether the image is represented by the given text, however, depending on the provided text, VE could quickly lead to complex scenarios. Recently, \cite{Hu2024CrackingTC} investigated the performances of Multimodal models on comics with contradictory narratives, where each comic consists of two panels that create a humorous contradiction. The paper introduces the YesBut benchmark and assesses Multimodal Models' limited capabilities in this reasoning task.


\subsection{Future Tasks}
In the context of Understanding, a number of Vision Language tasks could be of interest in comics, both for pertaining and fine-tuning. 

\vspace{2mm}
\subsubsection{Visual Dialog}~\\%
\textbf{Definition:} The task of Visual Dialogs (VisDial) corresponds to answering a specific question, having an image, a question about the image, and the previous dialog history about the image.

In comics, Visual dialog is an area still in its infancy and requires constructing a narrative dialogue based on visual cues within the comic panels. This task demands a comprehensive understanding of the storyline, character interactions, and visual symbolism to generate contextually relevant dialogues.

\vspace{2mm}
\subsubsection{Visual Reasoning}~\\%
\textbf{Definition:} Visual reasoning in comics extends beyond simple question answering to include the ability to interpret and analyze the visual content in depth. This task necessitates understanding the objects, their interactions, and the underlying narrative structure within the comic panels. It's a sophisticated blend of visual comprehension and logical deduction, aimed at uncovering deeper layers of meaning in the comics.

\section{Layer 5: Generation and Synthesis}
\label{sec:locu-layer5}
The last layer explores the creative frontier in comics analysis, where the synthesis and generation of comics from various media sources play a pivotal role.

\subsection{Generation}
\label{subsec:generation}
\begin{wrapfigure}{r}{0.4\textwidth}
    \centering
    \vspace{-0.2in}
    \includegraphics[width=0.4\textwidth]{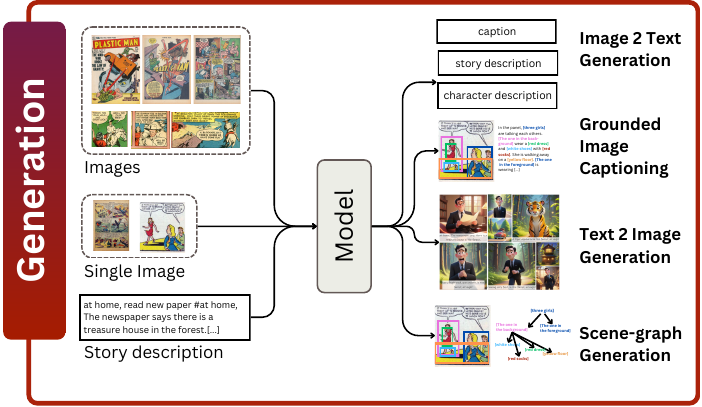}
    \vspace{-5mm}
    \caption{Illustration of \textit{Generation} tasks.}
    \vspace{-8mm}
    \label{ill:generation}
\end{wrapfigure}
The field of Generation in comics has seen lots of attraction on Manga style comics rather than American ones. Their popularity has attracted the research community towards creating comics from various media such as video, images, and text descriptions (Fig. \ref{ill:generation}).

\vspace{2mm}
\subsubsection{Comics generation from other media.}~\\%
\textbf{Definition:} The task of generating comics from other media comprehends an umbrella of tasks with various input types (video, audio, charts, and comics itself) whose output is a comic panel, a sequence of panels, a page, or a book.

One of the earliest attempts at comic generation from other media is the ``Comic Live Chat'' by Matsuda et al. \cite{matsuda_comic_2010}, which transformed video meetings into comics. This approach involved selecting keyframes and rendering the dialogues of meeting participants in a comic format. Tanapichet et al. \cite{tanapichet_automatic_2011} proposed a system for creating comic strips from cartoon animations using optical flow techniques, demonstrating a novel approach to narrative translation from one visual medium to another. Hoashi et al. \cite{hoashi_automatic_2011} worked on creating manga previews by detecting panels and text balloons, while in \cite{deng_research_2010, chang_caricaturation_2011} authors explored generating manga faces and caricatures from real images. Cao et al. \cite{cao_automatic_2012} and Herranz et al. \cite{herranz_scalable_2012} introduced methods for comic-like video summarization. These systems developed pipelines for layout proposal, image reconfiguration, and balloon creation, treating comic creation as a series of steps from keyframe selection to final panel layout. On the same task, to automatically create comics from video content, Jing et al. \cite{jing_contentaware_2015} pioneered video transformation using TV show subtitles and a speaker detection algorithm to generate comic layouts optimized through an energy-based metric. This approach was a significant step forward compared to earlier methods like \cite{wang_movie2comics_2012}.
Giving it more time, a groundbreaking approach by Pesko et al. \cite{pesko_comixify_2018} employed neural style transfer, creating visually compelling comic representations from video-sourced data\footnote{GitHub repository: \href{https://github.com/maciej3031/comixify}{https://github.com/maciej3031/comixify}}. While their startup has since become closed-sourced, their initial approach signifies a significant leap in comic generation technology. Yang et al. \cite{yang_automatic_2021} presented a system to generate comics from movies, employing a multi-page layout framework and emotion-aware balloon generation, showcasing the potential of comprehensive systems in comic generation. To extract keyframes from movies, they used GIST \cite{oliva_modeling_2001} similarity between two frames. To detect the speaker, they perform lip motion analysis and use that to associate balloons with talking characters. Finally, the movie frames are rendered in a comic-layout page after the adaption to comic style.
Moreover, in \cite{zhang_augmenting_2023} authors represent conversational videos to comics ``fixed-layout'' representation, developing a system that learns guidance ﬁeld which provides a prior prediction of the possible positions of word balloons while making the word balloons not overlap with other nonverbal information (e.g. hand gestures, visual clues in the background, etc). Their input is raw video, thus the text dialog is captured from audio and rendered in the location chosen from their method.
Comics have been used also in education \cite{suh_using_2021, lima_comics_2022, castro_developing_2023} to teach and explain in an easier way different concepts. An example of this is \cite{zhao_chartstory_2023} where authors proposed a system that crafts data stories from a collection of user-created charts, using a comic-style panel to imply the underlying sequence and logic of data-driven narratives.

\vspace{2mm}
\subsubsection{Image-to-Text generation}~\\%
\textbf{Definition:} Image-to-Text is a general umbrella term that covers tasks from captioning, to textbox prediction in comics, masking a specific part of the dialog.

Motivated by supporting the Blind or Low Vision community (BLV) - also known as People with Visual Impairements (PVI) - Ramaprasad et al. \cite{ramaprasad_comics_2023} developed a two-step method to create natural language descriptions of comic strips. This method first extracts information about panels, characters, and text using computer vision techniques, followed by the use of a multimodal large language model (MLLM) to generate descriptions.
This work stands as one of the few promoting MLLM specifically for comic strip captioning, despite using ``out of the box'' architectures like Grounding DINO \cite{liu_grounding_2023} to extract arbitrary elements, CLIP \cite{radford_learning_2021} to match character images with name-descriptions, and LLaVa \cite{liu_visual_2023} as MLLM.
Vivoli et al. \cite{vivoli_multimodal_2024} proposed an advanced version of the text-cloze task with image-to-text generation, increasing the challenge level. They adopted the VL-T5 model \cite{cho_unifying_2021} using two configurations with similar performances but different model sizes: with BLIP-2 visual backbone (1.3B) or a ResNet-50 fine-tuned with SimCLR on comics (0.2B). They surpassed previous state of the art in the standard ``text-cloze'' task, and propose benchmarks in the generation version of the task.
Authors in \cite{agrawal_multimodal_2023} presented a similar task but incorporated character descriptions and transcriptions, emphasizing a language model-centric approach. Guo et al. \cite{guo_m2c_2023} introduced the multi-modal manga complement task. This innovative task combines visual sequences from comic pages with corresponding text dialogues, challenging the model to complete the narrative appropriately. They propose an effective method with CLIP as Feature Encoding, a custom cross-attention module called Fine-grained Visual Prompt Generation, and a transformer-based encoder-decoder architecture called Dialog complement.
In a recent work \cite{sachdeva_manga_2024}, authors proposed a Relationformer-based architecture that generates dialog for manga pages. Their approach includes detection, matching, and sorting algorithms to produce an ordered dialog that aligns with the visual narrative.
However, all these works are limited by the dialog appearing in the comic, and in the context of describing a comic, not only the dialog is important, but also the scene description. Recent works, to fill this gap, have approached the task by exploring single panel captioning. 
In \cite{Rigaud2024TowardAC}, the authors explored prompt engineering with contextual information for closed-sourced Multimodal LLMs aiming at generating an accurate text description of a single panel and subsequently of the full story. They successfully demonstrated to retain many of the important details in the caption, but the exploration was limited to a couple of story panels. In \cite{Vivoli2024ComiCapAV}, authors took a step further proposing a new metric for comic panel captioning called ``attribute retaining metric'' (ARM) which assesses whether all the objects and attributes of the panel have been identified. They propose a pipeline with open-source Multimodal LLMs and zero-shot grounding models for generating dense and grounded captions for more than 1M panels.
However, in the ARM metric only the attributes are taken into consideration, which identifies the need to consider both locations and object/attribute names.
Finally, \cite{Hu2024CrackingTC} focuses on comics with contradictory narratives, where each comic consists of two panels that create a humorous contradiction, and introduces the ``YesBut'' benchmark, which comprises tasks of varying difficulty aimed at assessing MLLMs capabilities in recognizing and interpreting these comics.

\vspace{2mm}
\subsubsection{Text-to-Image generation}~\\%
\textbf{Definition:} The Text-to-Image task comprehends generating an image, or a sequence of images, from a text description.

In the field of Text-to-Image generation, early work by Jin et al. \cite{jin_automatic_2017} utilized GANs to generate anime-style character faces, demonstrating the potential of GANs for stylized image generation. Inoue et al. \cite{inoue_crossdomain_2018} advanced the field with domain transfer and pseudo-labeling techniques for image generation.
StoryGAN \cite{StoryGAN_2018} was a groundbreaking development in story visualization, employing a sequential conditional GAN with a context encoder to track story flow and a story-level discriminator. Later improvements included DuCo-StoryGAN \cite{maharana_improving_2021}, which used a dual learning framework for better semantic consistency, and VLStoryGAN \cite{VLStoryGAN_2021}, which incorporated recursive architectures to handle structured text inputs. Melistas et al. \cite{melistas_deep_2021} proposed a pipeline for generating synthetic graphic novels, using GPT-2 for text and StyleGAN2 for image synthesis, trained on a large manga collection. Proven-Bessel et al. \cite{proven-bessel_comicgan_2021} developed ComicGAN, a text-to-image GAN for generating single-panel comics in the style of Dilbert, while FastGAN \cite{liu_faster_2021}, enhanced for small datasets, was applied to manga faces \cite{hiruta_conditional_2022} with improved FID scores.
Yu et al. \cite{Yu2023ASO} introduced a multilingual text-to-image model for webtoon generation, showing the flexibility of GANs across languages. However, even fine-tuning on Korean MSCOCO resulted in abstract images that differed from authentic comics.
Diffusion models have also made strides in story visualization. StoryDALL-E \cite{maharana_storydalle_2022} used autoregressive Transformers for improved model tuning. Everaert et al. \cite{everaert_diffusion_2023} adapted Stable Diffusion for comic generation by tweaking the initial latent tensor, showing how diffusion models can be tailored for comic styles. Despite limitations in text generation, diffusion models successfully generated comic panels aligned with COMIC data distributions. The consistency of characters and style across multiple images, especially in storytelling contexts, remains a key challenge \cite{he_dreamstory_2024}.
Recent work by Jin et al. \cite{jin_generating_2023} leveraged large language models (LLMs) and fine-tuned Stable Diffusion to generate new manga content. Using ChatGPT to generate storylines and dialogues for a hypothetical continuation of the manga One Piece, they fine-tuned Stable Diffusion with LoRA and ControlNet to match the layout, style, and aesthetic of the original manga. The final product included LLM-generated dialogue integrated into comic panels.
The subfield of story visualization within text-to-image generation is rapidly evolving, driven by the integration of powerful LLMs and new image synthesis techniques. Key challenges include maintaining character and style consistency across multiple panels and developing unified benchmarks to track advancements across the field.

\subsection{Synthesis}
\begin{wrapfigure}{r}{0.4\textwidth}
    \centering
    \vspace{-0.2in}
    \includegraphics[width=0.4\textwidth]{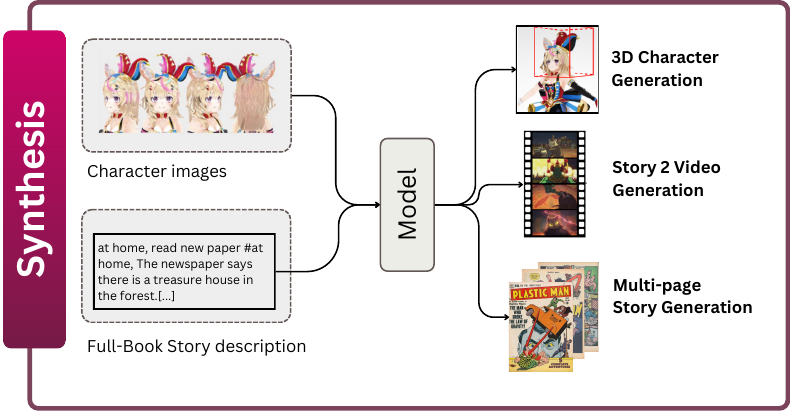}
    \vspace{-5mm}
    \caption{Illustration of \textit{Synthesis} tasks.}
    \vspace{-5mm}
    \label{ill:synthesis}
\end{wrapfigure}

The group of Synthesis tasks, compared to the ``generation'' one, refers to the creation of more complex, structured, and often temporally extended outputs, such as creating a full graphic novel from a complex storyline or merging multiple elements (characters, scenes, dialogues) in a consistent way across multiple pages (Fig. \ref{ill:synthesis}).

\vspace{2mm}
\subsubsection{3D generation from images}~\\%
\textbf{Definition:} The task of \textit{generating 3D models from 2D comic illustrations} presents unique challenges. Unlike conventional portrait illustrations, comics often involve stylized and exaggerated features, adding complexity to this already intricate process.

While there are advancements in using 3D character assets for tasks like pose estimation \cite{kim_ugatit_2020}, re-targetting \cite{khungurn_pkhungurn_2022,kim_animeceleb_2022}, and character reposing \cite{lin_collaborative_2023}, the field lacks scalable training resources due to the unavailability of suitable 3D character assets. In light of these issues, Chen et al. \cite{chen_panic3d_2023} addressed this gap by formalizing the stylized reconstruction task with the introduction of the \textit{AnimeRecon} benchmark and the \textit{Vroid} dataset. The Vroid dataset, in particular, provides a wealth of 3D assets for scalable training in this domain. A significant part of their work involves solving the challenge of contour removal from illustrations, which is a crucial step in achieving accurate 3D reconstruction from stylized comic images.
While there are numerous works on 3D reconstruction from line drawings in general \cite{lun_3d_2017, sanghi_sketchashape_2023}, the specific application in comics is underexplored, likely due to the absence of suitable datasets and ground truth references for comics.
It is worth noting that the exploration of 3D reconstruction from comic images holds great potential. It not only enriches the visual experience of comics but also opens up possibilities for animations, interactive media, and virtual reality applications within the comic domain.

\vspace{2mm}
\subsubsection{Video generation}~\\%
\textbf{Definition:} Video generation in comic refers to the \textit{synthesis of video content from static comic panels}. This process represents a significant leap in merging traditional comic art with dynamic multimedia.

Recent advancements in video generation from both images \cite{yan_motionconditioned_2023, wang_dreamvideo_2023} and text \cite{gupta_photorealistic_2023} have made significant strides, though defining accurate metrics and benchmarks remains a challenge \cite{fan_aigcbench_2024}. These developments are rapidly evolving within the broader Vision and Language community.
However, in the domain of comics, video generation is still in its early stages. One of the first attempts came from Cao et al. \cite{cao_dynamic_2017}, who introduced a framework to animate manga panels. Their method classifies panels based on motion and emotion, using this information to animate characters and backgrounds. The resulting video includes both intra-panel animations and transitions between panels. Later, Gupta et al. \cite{gupta_c2vnet_2021} expanded on this with C2VNet, which generates panel-audio videos by cropping panels and adding a highlighting effect to speech balloons to simulate character dialogue. They also introduced IMCDB, a dataset of annotated Indian Mythological Comics in English, to support research in this field.
While these efforts represent initial steps, they are still far from the capabilities seen in Vision-Language video generation, such as producing fully animated video clips with sound from a single image. The integration of comics with video and audio elements offers exciting potential for enhancing storytelling and audience engagement.
Future research could pave the way for interactive and immersive comic experiences, blending traditional art with modern digital technologies.

\vspace{2mm}
\subsubsection{Narrative-based Comic Generation}~\\%
\textbf{Definition:} The task involves generating a comic (whether a panel, strip, or page) from a plain text description. The term ``Narrative-based'' highlights that the text not only describes the scene but also conveys the story's narration.

The pioneering efforts in the realm of narrative-based comic generation commenced with the introduction of StoryGAN by Li et al. \cite{li_storygan_2019}, which marked a significant advancement in generating image sequences from storylines using a sequential conditional GAN framework. This innovation incorporated a Context Encoder, which dynamically tracked the story flow, laying the groundwork for future developments in this area.
Building on this foundation, Maharana et al. \cite{maharana_improving_2021} enhanced visual quality and coherence by incorporating a dual learning framework that utilized video captioning, a copy-transform mechanism, and MART-based transformers. This approach significantly improved the semantic alignment between the story and generated images, thus enriching the story visualization process.
In terms of character coherence and continuity, Chen et al. \cite{chen_charactercentric_2022} focused on maintaining consistent character portrayal throughout the narrative. They adapted Vector-Quantized Variational Autoencoders (VQ-VAE) with a text-to-visual-token architecture, ensuring character coherence in the visual narratives. Furthering this direction, Maharana et al. \cite{maharana_storydalle_2022} introduced the concept of story continuation, where the generated visual story is conditioned on a source image, allowing narratives to incorporate new characters more fluidly. They also introduced the DiDeMoSV dataset, which provided a platform for exploring these complex narrative structures.
Recent advancements in story visualization are exemplified by Pan et al. \cite{pan_synthesizing_2022}, with presenting AR-LDM, a latent diffusion model that significantly raised the standards for visual quality in natural image datasets like VIST. This model was auto-regressively conditioned on history captions and generated images, capturing complex interactions between frames. Rahman et al. \cite{rahman_makeastory_2023} introduced a framework with a visual memory module that implicitly captured actor and background context, generating frames that were not only of high visual quality but also consistent with the story. Peng et al. \cite{peng_personalized_2023} proposed PCSG, a diffusion-based text-to-image synthesis framework with controllable plugins for character consistency, scene layout specification, and character pose specification, further enhancing the personalized aspects of story visualization.
These emerging trends and future directions in narrative-based comic generation showcase a notable shift towards more sophisticated, context-aware, and visually coherent storytelling techniques, potentially leading to automated comic creation tools that adapt to various narrative styles and complexities. 

\subsection{Future Tasks}

\vspace{2mm}
\subsubsection{Comics to Scene graph}~\\%
\textbf{Definition:} Transforming comics into scene graphs is an emerging area of research. This task involves dissecting comic panels to identify and represent the relationships between different elements within the panel in a graph format. 

The last work that tackles this task is \cite{sachdeva_manga_2024}, which employs a Relationformer architecture \cite{shit_relationformer_2022}, explicitly designed for image-to-graph tasks. However, the authors convert the full page into different types of nodes (panels, text, characters) and only consider two types of arches: characters-characters (re-identification) and text-characters (speaker identification). The inherently complexity of the comics, such as the elements in the image (e.g. the background and foreground objects) have not been considered yet. The task has yet to be explored to fine-grained graph creation from single panels, full pages, or a book.

\section{CONCLUSION}\label{sec:conclusion}

In reviewing the diverse and innovative efforts in the realm of comic research and technology, it becomes evident that this field is in a state of dynamic and exciting evolution. The array of tasks and applications that researchers have embarked upon reflects not only the complexity of comics as a medium but also their immense potential as a bridge between artistic expression and technological advancement.
We have covered numerous works and categorized them into distinct layers based on a proposed taxonomy. We have aimed to distill the key topics, methods, challenges, and emerging directions that can shape future research in \comicsunderstanding.
In the future, we will continue to monitor advancements in this field, and we will update our findings in the following repository: \url{https://github.com/emanuelevivoli/awesome-comics-understanding}.


\bibliographystyle{ACM-Reference-Format}
\bibliography{main}

\appendix




\end{document}